\documentclass[twoside]{article}

\usepackage[hidelinks]{hyperref}
\usepackage{breakcites}   

\usepackage{amssymb}            
\usepackage{amsthm}
\usepackage{mathtools}          
\usepackage{mathrsfs}           
\usepackage{graphicx}           
\usepackage{subcaption}         
\usepackage[space]{grffile}     
\usepackage{url}                
\usepackage{lipsum}             
\usepackage{algorithm}
\usepackage{float}
\usepackage{booktabs}
\usepackage{algorithmic}
\usepackage{tikz}
\usetikzlibrary{positioning}

\newtheorem{definition}{Definition}

\usepackage{wrapfig}
\usepackage[accepted]{aistats2026}
\usepackage[round]{natbib}

\begin{document}
\runningtitle{Bayesian Inverse Transition Learning}
\frenchspacing

%

%

\twocolumn[

\aistatstitle{Bayesian Inverse Transition Learning: \\Learning Dynamics From Near-Optimal Trajectories}

\aistatsauthor{Leo Benac \And Abhishek Sharma \And Sonali Parbhoo \And Finale Doshi-Velez}

\aistatsaddress{Harvard University \And Harvard University \And Imperial College London \And Harvard University}


]
\begin{abstract}
  We consider the problem of estimating the transition dynamics $T^*$ from near-optimal expert trajectories in the context of offline model-based reinforcement learning. We develop a novel constraint-based method, Inverse Transition Learning, that treats the limited coverage of the expert trajectories as a \emph{feature}: we use the fact that the expert is near-optimal to inform our estimate of $T^*$. We integrate our constraints into a Bayesian approach. Across both synthetic environments and real healthcare scenarios like Intensive Care Unit (ICU) patient management in hypotension, we demonstrate not only significant improvements in decision-making, but that our posterior can inform when transfer will be successful. 
\end{abstract}

\section{INTRODUCTION}
In traditional planning scenarios, the rewards \(R\) and transition dynamics \(T^*\) of the environment are known, and the goal is to compute the optimal policy \(\pi^*\) that maximizes long-term returns. However, in many real-world situations, the transition dynamics \(T^*\) are unknown. Model-Based Reinforcement Learning (MBRL) addresses this by first learning \(T^*\) and then performing planning, which improves data efficiency and enables counterfactual reasoning \citep{sutton2018reinforcement, ghavamzadeh2015bayesian, kidambi2020morel, yu2021combo, lee2021representation, poupart2006analytic, ha2018recurrent, oh2015action, buesing2018woulda}.

This paper focuses on learning the true dynamics \(T^*\) in offline settings using batch data generated by a near-optimal expert. This is a common setting in fields like healthcare and education, where one can presume that the behavior policy is imperfect but generally reasonable. Learning \(T^*\) from observational data is challenging due to low coverage of the state-action space. Without the ability to interact with the environment, it is crucial to utilize the limited data effectively.
We leverage the knowledge that the trajectories are near-optimal to better estimate $T^*$. 
Consider the simple 2-state MDP shown below.
The blue path shows the observed optimal behavior leading to the goal state \(s_{\text{goal}}\) from \(s_1\) after action \(a_1\).  The dashed lines represent the alternative hypothetical paths following an unobserved action $a_2$.  The fact that the expert chose action $a_1$ implies there must be a higher likelihood of reaching \(s_{\text{goal}}\) via \(a_1\) than \(a_2\) (i.e., \(T^*(s_{\text{goal}}| s_1, a_1) > T^*(s_{\text{goal}}| s_1, a_2)\).

\begin{wrapfigure}[7]{r}{0.44\linewidth}
    \vspace{-0.8\baselineskip}
  \label{easyexample}
  \centering
  \begin{tikzpicture}[node distance=1.38cm, auto, every node/.style={scale=0.9}]
      \node (s1) [text=blue] {$s_1$};
      \node (sgoal) [right=of s1] {$s_{\text{goal}}$};
      \draw[->, thick, blue] (s1) to[bend left=30] node[above] {$a_1$} (sgoal);
      \draw[->, dashed] (s1) to[bend right=30] node[below] {$a_2$} (sgoal);
      \path (s1) edge [loop left, dashed] node[left] {$a_2$} (s1);
      \node [align=left, below=of s1, yshift=0.7cm, xshift=0.5cm] (legend) {
          \textcolor{blue}{\textbf{Observed expert}}: $a_1$\\
          \textbf{Unobserved}: $a_2$
      };
  \end{tikzpicture}
  \label{fig:easyexample}
  \vspace{-2\baselineskip}
\end{wrapfigure}

A popular definition of near-optimality models an expert's actions as being proportional to the action's values, as presented in Maximum Causal Entropy (MCE) Inverse RL \citep{ziebart2010modeling}.

In clinical settings, clinicians often prefer to administer treatments from drug families that have similar, reasonably good effects, without assigning a ranking. Conversely, they avoid poorly performing treatments, also without assigning a ranking. Instead of selecting a single best action, which may be influenced by data collection or artifacts, it is often better to focus on sets of actions that are nearly equivalent in performance \citep{tang2020clinician} and identify which poor behaviors to avoid \citep{fatemi2021medical,tang2022leveraging}.

The most closely related works to ours \citep{herman2016inverse, reddy2018you} rely on the Maximum Causal Entropy (MCE) approach. While modeling near-optimal expert behavior this way is reasonable, our approach shows better performance in clinical settings. Moreover, solving for $T^*$ in the MCE framework requires a gradient-based alternating optimization, which is computationally intensive and prone to local optima. Finally, the constraints in these methods are soft, so their learned dynamics lack guarantees.

To address these limitations, we introduce a novel approach called \emph{Inverse Transition Learning} (ITL), based on hard constraints on the true dynamics $T^*$ and the distinction between two groups of actions. Our constraints ensure that executed actions have higher value than actions not taken. If multiple actions are taken in the same state, they are constrained to be close in value. These thresholds are governed explicitly by a transparent, human-understandable parameter, rather than implicitly through the entropy of an action-value distribution. They can also be solved deterministically via a quadratic program (e.g., using CVXPY \citep{diamond2016cvxpy}), avoiding the drawbacks of gradient-based optimization. This yields an optimization procedure that provides guarantees for the estimated dynamics and converges significantly faster.

The fact that the trajectories are near-optimal provides some information about what transition dynamics $T^*$ are feasible, but limited coverage in the batch setting means that some uncertainty will still remain.  Thus, we extend our approach to the Bayesian model-based setting \citep{dearden2013model}.  We develop an efficient approach to posterior estimation that includes our constraints. Our approach narrows the gap to the true dynamics \(T^*\) compared to baseline Bayesian MBRL methods. Additionally, given a new reward function, we demonstrate how keeping a posterior over the true dynamics $T^*$ can be used to predict when the transfer will be successful. Our main contributions are:
\begin{itemize}
   \item We propose a fast, transparent, and more reliable method to learn a point estimate of the true unknown dynamics \(T^*\) by leveraging expert demonstrations. Our method avoids gradient-based optimization, does not rely on the MCE modeling approach, and enforces constraint guarantees on the learned dynamics.
    \item We incorporate Bayesian inference to learn a posterior distribution over \(T^*\). The posterior offers the same constraints guarantees for each sample, allows us to quantify the uncertainty over actions, and can be used to predict on which reward functions we except to perform well.
\end{itemize}
Although our method is designed for tabular MDPs, we demonstrate its effectiveness in continuous domains as well. We tested our approach in a range of environments, including a synthetic Gridworld, 10 different Randomworld environments, and a real-world healthcare setting. These experiments covered various levels of data coverage and expert optimality. Our results show that our method performs well across different metrics, consistently outpacing baseline methods in both synthetic and real-world scenarios.
\section{RELATED WORKS}
\vspace{-0.1cm}
\textbf{Model-Based Reinforcement Learning.}
We explore the utilization of forward models, which predict the subsequent state \(s'\) from the current state \(s\) and action \(a\) (\((s, a) \rightarrow s'\)), as defined in \citep{moerland2023model}. These models are central for understanding action-induced state transitions. In contrast, backward models identify potential antecedents to states (\(s' \rightarrow (s, a)\)) and are used for backward planning \citep{moore1993prioritized}. Similarly, inverse models compute actions required to transition between states \((s, s' \rightarrow a)\), beneficial in RRT planning \citep{lavalle1998rapidly}. Additionally, non-parametric methods like replay buffers enable precise estimations \citep{lin1992self,vanseijen2015deeper, van2019use}, and approximation methods, such as Gaussian processes, offer alternatives \citep{wang2005gaussian,deisenroth2011pilco}. In discrete MDPs like ours, tabular maximum likelihood estimation models, noted as \(T^{MLE}\), represent the state of the art \citep{sutton1991dyna}.
\vspace{-0.2cm}

\paragraph{Bayesian Model-Based Reinforcement Learning}
Bayesian MBRL integrates uncertainty into model learning, as detailed in foundational works \citep{ghavamzadeh2015bayesian, ross2008model, dearden2013model}. Unlike earlier approaches, such as \citep{poupart2008model}, which address partially observable settings in discrete factored domains by updating beliefs with new data, our method applies Bayesian MBRL in an offline setting. We infer posterior distributions over transition dynamics solely from expert knowledge. This approach combines prior knowledge with batch data to develop models that yield reliable policies \citep{guo2022model}. In contrast, studies like \citep{zhang2020invariant,zhang2020learning} focus on learning invariant representations for control without explicitly modeling transition dynamics.

\textbf{Learning from Demonstrations.}
\emph{Imitation Learning} (IL) learns policies directly from expert demonstrations via per-timestep training \citep{ross2010no}, iterative expert feedback (DAgger) \citep{ross2011reduction}, or imposing linear constraints during policy optimization (APID) \citep{kim2013learning}. In contrast, \emph{Inverse Reinforcement Learning} (IRL) infers the reward $R$ from dynamics $T$ and demonstrations to facilitate transferability \citep{ng2000algorithms}, using methods like Maximum Margin and Maximum Entropy to robustly mimic experts \citep{abbeel2004apprenticeship, ziebart2008maximum, scobee2019maximum}. To address $R$'s non-identifiability, Bayesian IRL learns a distribution over $R$ \citep{ramachandran2007bayesian}. Distinct from these policy and reward-focused approaches, our \emph{Inverse Transition Learning} framework utilizes expert demonstrations to directly refine the estimation of the true dynamics $T^*$, contrasting with standard baselines that rely solely on maximum likelihood.


\section{PRELIMINARIES}
\textbf{Markov Decision Processes (MDPs).} An MDP $\mathcal{M}$ can be represented as a tuple $\mathcal{M} = \{\mathcal{S}, \mathcal{A}, T^*, \gamma, R\}$, where $\mathcal{S}$ is the state space, $\mathcal{A}$ the action space, $T^*$ is the true dynamics of the environment, $\gamma$ is the discount factor, and $R$ is a bounded reward function. In planning, the goal is to find the best policy $\pi^*$ corresponding to an MDP $\mathcal{M}$. In this paper, we focus on tabular MDPs (discrete state and action spaces). Discretization of continuous clinical variables can be achieved using domain knowledge and is a common and effective practice in healthcare RL applications. This approach aligns naturally with clinical decision-making, where physicians often categorize patient states and treatment options into discrete, clinically meaningful groups. For example, \citep{choudhary2024icu} use the same healthcare dataset as us (MIMIC \citep{johnson2020mimic}) to create an RL simulator by discretizing both state and action spaces. Such discretization not only reflects clinical practice but also makes the learned dynamics more interpretable in low-data settings.

\textbf{Value Functions.}
\looseness=-1
For a policy $\pi$, the state–value function $V^\pi(s)$ is the expected discounted return starting from state $s$, and the action–value function $Q^\pi(s,a)$ is the expected discounted return starting from state $s$ and taking action $a$.

\textbf{Notation.}
\looseness=-1
Let $V^{\pi}$ denote the vector of values $V^{\pi}(s)$. We use shorthand
$R_a$ , $Q_a$, and $T_a$ to represent the vectors $R(\cdot, a)$, $Q(\cdot, a)$ and the matrix $T(\cdot \mid \cdot, a)$. We also use shorthand $R_{\pi}, Q_{\pi}$, and $T_{\pi}$ to represent the vectors $\mathbb{E}_{a \sim \pi}[R_a]$.
$\mathbb{E}_{a \sim \pi}[Q_a]$ and the matrix $\mathbb{E}_{a \sim \pi}[T_a]$. For dynamics $T$, we let $\pi^*(T)$ and $Q^*(T)$ denote the corresponding optimal policy and Q values function respectively with respect to dynamics $T$. In tabular settings, the value functions $V^{\pi}$ and $Q^{\pi}_a$ can be calculated directly through a closed-form solution:   
\begin{gather}
    V^\pi = R_\pi + \gamma T_\pi V^\pi =  (I - \gamma T_\pi) ^{-1} R_\pi \\ Q^{\pi}_a =   R_a + \gamma T_a  (I - \gamma T_\pi) ^{-1} R_\pi.
\label{eq: closed form solutions values }
\end{gather}

\textbf{Methods for Estimating the Transition Dynamics $T^*$.} In tabular settings, a common method to estimate the true transition dynamics $T^*$ is the Maximum Likelihood Estimate (MLE), denoted as $T^{MLE}$ (e.g., \citep{barto1995learning,kim2023model,ornik2021learning}). The number of times the tuple (s, a, s') is observed is denoted by $N_{s, a, s'}$. We apply Laplace smoothing, represented by $\delta$, to address issues such as zero occurrences in the count data due to low coverage in our batch data $\mathcal{D}$. The $T^{MLE}$ is defined as:
\begin{align}
    \label{MLE baseline}
    T^{MLE}(s'|s, a) = \frac{N_{s, a, s'} + \delta}{\sum_{s'} (N_{s, a, s'} + \delta)}
\end{align}

In offline tabular settings, \citep{herman2016inverse} leverage near-optimal batch data to infer dynamics by reusing the MCE IRL framework \citep{ziebart2010modeling}. We include this method in our baseline and refer to it as MCE (or $T^{MCE}$) moving forward. Their method consist of an interative procedure between taking one gradient step of the dynamics parameters $\theta$ with respect to the loss in equation \ref{baseline equation} and then performing soft-Value Iteration, where Q is the soft Q function.
\begin{equation}
\label{baseline equation}
\sum_{(s,a,s') \in \mathcal{D}} 
\left[
  \log \frac{\exp(Q_{\theta}(s,a))}{\sum_{a'} \exp(Q_{\theta}(s,a'))}
  + \log T_{\theta}(s' \mid s,a)
\right].
\end{equation}
To model the uncertainty of transitions \( T^* \) probabilistically, \citep{ghavamzadeh2015bayesian} uses a Multinomial likelihood in conjunction with a Dirichlet prior. This combination yields the posterior \( P(T^*| \mathcal{D}) \), which estimates the transition probabilities given data \( \mathcal{D} \).
\begin{equation}
\label{eq: ps equation}
{\small
\begin{aligned}
P(T^*(\cdot \mid s, a) \mid \mathcal{D} ) &= \text{Dir}(\mathbf{N}_{s,a} + \mathbf{\delta} \mid s, a) \\
&\propto \text{Multinomial}(\mathbf{N}_{s,a} \mid s, a) \cdot \text{Dir}(\mathbf{\delta} \mid s, a)
\end{aligned}
}
\end{equation}
Note that \( T^{MLE} \) represents the mean of the distribution \( P(T^*| \mathcal{D}) \). However, this posterior distribution is not as tight as it could be because it does not account for the near-optimality of expert trajectories. To refine this, we utilize expert signals to develop a tighter posterior \( P(T^*| \mathcal{D}) \), enhancing the estimation of the transition dynamics. To the best of our knowledge, we are the first to infer a distribution over the dynamics using expert demonstrations. Furthermore, we propose a method distinct from \citep{herman2016inverse} to derive a point estimate of \( T^* \). This involves formulating constraints based on expert behaviors, offering a more accurate approach to modeling the true dynamics.

\section{ESTIMATING TRANSITION DYNAMICS WITH $\epsilon$-OPTIMAL EXPERT}
\label{definitions}
This section introduces key definitions to relate the expert's optimality to the true dynamics $T^*$. 

\begin{definition}[{\emph{$\epsilon$-ball}}]
For any state $s$ and transition dynamics $T$, an action $a$ is in the $\epsilon$-ball $\epsilon(s; T)$ if it is $\epsilon$-close to the optimal action according to its optimal Q-function $Q^*(.,.; T)$ with respect to dynamics $T$:
$$a \in \epsilon(s; T) \iff  \max_{a'} Q^{*}(s, a'; T) - Q^{*}(s, a; T) \leq \epsilon$$
\end{definition}
For each state $s$, $\epsilon(s; T^*)$ contains the actions that are near-optimal with respect to the true dynamics $T^*$. 
\begin{definition}[{\emph{$\epsilon$-optimality}}]
\label{eps optimal expert}
A policy $\pi_{\epsilon}(.| s; T^*)$ is $\epsilon$-optimal with respect to the true unknown transition dynamics $T^*$ if it exclusively selects actions from the $\epsilon$-ball $\epsilon(s; T^*)$ for all states $s$:
$$\pi_{\epsilon}(a|s; T^*) > 0 \iff a \in \epsilon(s; T^*), \quad \forall s \in \mathcal{S}$$
\end{definition}
Such a policy $\pi_{\epsilon}(.| s; T^*)$ can make a mistake of at most $\epsilon$ at each time step relative to $Q^*(.,.;T^*)$. We assume our demonstrations are generated by such policy.

\begin{definition}[{\emph{$\epsilon$-ball property}}]
\label{eps ball property}
 Dynamics $T$ satisfy the $\epsilon$-ball property for state $s$ if $\epsilon(s; T) = \epsilon(s; T^*)$ and \textbf{every} action $a' \notin \epsilon(s; T)$ is at least $\epsilon$-away from all actions $a \in \epsilon(s;T)$ with respect to $Q^*(s,.; T)$:
\begin{equation}
\label{eps ball property eq}
{\small
\begin{aligned}
&\epsilon(s; T) = \epsilon(s; T^*) \quad \text{and} \\
&Q^*(s, a; T) - Q^*(s, a'; T) \geq \epsilon, \\
&\qquad \forall a \in \epsilon(s;T), \, \forall a' \notin \epsilon(s; T)
\end{aligned}
}
\end{equation}
\end{definition}
We assume the true dynamics $T^*$ satisfy the $\epsilon$-ball property for each state $s$, splitting actions into two groups: near-optimal and suboptimal. Actions inside $\epsilon(s;T^*)$ are \textbf{valid}, and those outside $\epsilon(s;T^*)$ \textbf{invalid}. Our goal is to learn dynamics $T$ that satisfy this property for every state $s$. Such dynamics can distinguish near-optimal actions from suboptimal ones. 

\begin{definition}[{\emph{Deterministic/stochastic-policy state}}]
\label{policy states}
A deterministic-policy state occurs when the $\epsilon$-optimal expert $\pi_{\epsilon}(.|s; T^*)$ selects a single action in state $s$ (meaning the expert is deterministically optimal). A stochastic-policy state occurs when the expert selects multiple actions in such state:
$$\text{Deterministic-policy state:} \quad |\epsilon(s; T^*)| = 1$$
$$\text{Stochastic-policy state:} \quad |\epsilon(s; T^*)| > 1$$
\end{definition}
Deterministic-policy states are akin to conditions well-understood by clinicians who are certain of the best treatment, whereas stochastic-policy states resemble conditions where multiple reasonable treatments exist without clear superiority.

\paragraph{Problem Setting}
We assume we are given the MDP \(\mathcal{M}\) except for the true transition dynamics \(T^*\) (i.e., \(\mathcal{M} \setminus \{T^*\}\)). We also have access to batch data \(\mathcal{D} = \{(s_i, a_i, s'_i)\}_{i=1}^N\), where the data is assumed to have been generated by the behavior policy $\pi_{\epsilon}(.| s; T^*)$, which is $\epsilon$-optimal with respect to the true dynamics $T^*$. Note that when \(\epsilon = 0\), the expert \(\pi_{\epsilon}(. | .; T^*)\) is fully optimal, and hence we would only encounter deterministic-policy states.


\paragraph{Why Access to the Reward Function is Reasonable}
\looseness=-1
Unlike transition dynamics, which describe the complex physiological response of the human body to treatments and are difficult to infer even with domain knowledge, the reward function in healthcare applications is often more accessible. In the case of hypotension management, clinical expertise can reasonably define rewards that distinguish good and bad outcomes, such as stabilizing blood pressure or preventing adverse events. Therefore, in our setup, it is natural to assume access to the reward function while focusing on learning the transition dynamics \(T^*\). However, we recognize that in other domains, such as robotics, specifying a reward function can be challenging. For such settings, we provide an extension of our algorithm in Appendix \ref{sec: joint_dynamics_rewards_learning} that jointly learns the transition dynamics and the reward function simultaneously. Our results demonstrate that this joint learning approach achieves performance comparable to when the true reward function is provided.

\subsection{Constraints on Transition Dynamics Given \(\pi_{\epsilon}(. |.; T^*)\)}
\label{sec:constraints_w_expert}
In this section, we develop a set of constraints for learning transition dynamics \(T\) motivated by the \(\epsilon\)-optimal assumption we have on the expert (See Definition \ref{eps optimal expert}). We denote these constraints on the dynamics $T$ collectively as \textbf{Constraints(\(T\))}, which consist of two types:
\vspace{-0.5cm}
\paragraph{Constraint 1(\(T\)): Separation Between Valid and Invalid Actions}
\looseness=-1
For every valid action \(a \in \epsilon(s; T^*)\) and every invalid action \(a' \notin \epsilon(s; T^*)\), \(T\) must satisfy: 
\begin{align}
    Q^{*}(s, a; T) - Q^{*}(s, a'; T) \geq \epsilon
    \label{eq: constraints1}
\end{align}
Expanding the Q-values in terms of \(T\):
\begin{align}
    R(s, a) - R(s, a') + \gamma \big(T(\cdot \mid s, a) \nonumber \\
    - T(\cdot \mid s, a')\big)^\top (I - \gamma T_{\pi^*(T)})^{-1} R_{\pi^*(T)} \geq \epsilon
    \label{eq: constraints1_expanded}
\end{align}
where \(\pi^*(T)\) denotes the optimal policy induced by dynamics \(T\). This constraint ensures that for every state \(s\), \emph{all} invalid actions are at least \(\epsilon\)-away from \emph{all} valid actions with respect to \(Q^{*}(s, \cdot; T)\).

\paragraph{Constraint 2(\(T\)): \(\epsilon\)-Closeness Among Valid Actions}
\looseness=-1
For each pair of valid actions \((a, a') \in \epsilon(s; T^*) \times \epsilon(s; T^*)\) where \(a \neq a'\), \(T\) must satisfy:
\begin{align}
    \left|Q^{*}(s, a; T) - Q^{*}(s, a'; T) \right| \leq \epsilon
    \label{eq: constraints2}
\end{align}
Expanding the Q-values in terms of \(T\):
\begin{align}
&\left| R(s,a) - R(s,a') 
   + \gamma \big(T(\cdot \mid s,a) - T(\cdot \mid s,a')\big)^\top \right. \nonumber \\
&\quad \left. (I - \gamma T_{\pi^*(T)})^{-1} R_{\pi^*(T)} \right| \leq \epsilon
\label{eq:constraints2_expanded}
\end{align}
This constraint ensures that for each state $s$, valid actions remain $\epsilon$-close with respect to $Q^*(s,\cdot;T)$.

\textbf{Enforcing the \(\epsilon\)-ball property:} Together, these constraints enforce the \(\epsilon\)-ball property (Definition \ref{eps ball property}) for the learned dynamics \(T\). \textbf{Constraint 2($T$)} ensures that all valid actions remain \(\epsilon\)-close to each other, while \textbf{Constraint 1($T$)} ensures that all invalid actions stay at least \(\epsilon\)-away from valid actions with respect to dynamics \(T\). This is exactly what the \(\epsilon\)-ball property requires. Hence, if dynamics \(T\) satisfy both sets of constraints, then it satisfies the \(\epsilon\)-ball property.

\subsection{Constraints on Transition Dynamics in practice}
\label{sec:constraints_w/o_expert}
While the true dynamics $T^*$ and sets $\epsilon(s;T^*)$ are unknown, we approximate them by collecting all actions $a$ taken in state $s$ from the dataset $\mathcal{D}$, denoted $\hat{\epsilon}(s;T^*)$. This relies on the assumption that $\mathcal{D}$ was generated by an $\epsilon$-optimal expert under the true dynamics $T^*$.

In practice, we do not have direct access to the expert policy $\pi_{\epsilon}(\cdot \mid s;T^*)$. Instead, we rely on the dataset $\mathcal{D}$, which may not cover all (state, action) pairs. We therefore define an estimated policy $\hat{\pi}_{\epsilon}(\cdot \mid s;T^*)$: for states $s \in \mathcal{D}$ it assigns equal probability to all observed actions, while for $s \notin \mathcal{D}$ it selects the best action under the proposed transition model $T$ (using $\pi^*(T)$). This policy is then used to define the constraints in Algorithm \ref{algo_HMC} $\&$ ~\ref{algo_ITL}, which integrates these components to estimate the true but unknown dynamics $T^*$.

\section{METHODOLOGY}
In this section, we describe how, given (near-)optimal data, our constraints can be applied to infer both a posterior distribution over the true dynamics $T^*$ and a point estimate. These are referred to as \emph{Bayesian Inverse Transition Learning} (BITL) and \emph{Inverse Transition Learning} (ITL), respectively.

\textbf{Bayesian Inverse Transition Learning}
Our constraints (Inequalities \ref{eq: constraints1} and \ref{eq: constraints2}) lead to an underdetermined problem with infinitely many solutions. Instead of introducing a loss function as the baseline MCE \citep{herman2016inverse}, we infer a posterior distribution on the true transition dynamics \(T^*\) to quantify such uncertainty. We introduce a sampling-based technique to infer this distribution, denoted as \(P_{\epsilon}(T^*| \mathcal{D})\), assuming data is generated by an \(\epsilon\)-optimal expert. We use our constraints to ensure each sample satisfies the \(\epsilon\)-ball property (See Definition \ref{eps ball property}) for each state.

\textbf{Naive Approach: Rejection Sampling}
The complexity of deriving a posterior from expert trajectories with many constraints precludes an analytic solution. An initial attempt might involve drawing samples from the simpler \(P(T^*| \mathcal{D})\) and rejecting those that fail to meet our constraints. This turns out to be very inefficient and rejects nearly all samples.


\textbf{HMC with Reflection}
To enhance sampling efficiency within our constrained, high-dimensional space, we employ Hamiltonian Monte Carlo (HMC) with reflection \citep{betancourt2011nested}. The log-likelihood function of our target distribution takes the following form: $\log P(T^* \mid \mathcal{D}) = \sum_{s, a} \log \text{Dir}( \mathbf{N}_{s,a} + \delta \mid s, a) $.
Applying standard HMC to Dirichlet distributions is challenging due to their simplex constraints. To address this, we use the invertible transformations from \citep{betancourt2012cruising}, combined with a logit transformation, to map the constrained transition dynamics $T$ into an unconstrained position variable $w$. 

Our sampling procedure (Algorithm \ref{algo_HMC}) utilizes first-order leapfrog integration. Let $m$ denote the momentum, $\alpha$ the step size, $\nabla E(w)$ the energy gradient, and $L$ the number of integration steps. During each spatial step, we transform $w$ back to $T^{(i)}$ to compute the optimal policy $\pi^*(T^{(i)})$ and check \textbf{Constraints 1 \& 2} (Inequalities \ref{eq: constraints1} \& \ref{eq: constraints2}). If a constraint $c_{s,a,a'}(T^{(i)})$ is violated, the trajectory ``bounces'' by reflecting the momentum $m$ off the constraint's normal vector $\hat{n}$.

\begin{algorithm}[htb!]
\caption{HMC with reflection for Bayesian ITL}
\label{algo_HMC}
\begin{algorithmic}[1]
    \STATE $m \gets m - \frac{1}{2}\alpha \nabla E(w)$ \hfill \textit{// First momentum half step}
    \FOR{$i = 0$ to $L$}
        \STATE $w \gets w + \alpha m$ \hfill \textit{// Full spatial step}
        \STATE $T^{(i)} \gets \text{w\_to\_T}(w)$
        \STATE Compute $\pi^*(T^{(i)})$ 
        \IF{all \textbf{constraints}($T^{(i)}$) are satisfied}
            \STATE $m \gets m - \alpha \nabla E(w)$ \hfill \textit{// Full momentum step}
        \ELSE 
            \STATE $\hat{n} \gets \nabla c_{s, a, a'}(T^{(i)}) / \| \nabla c_{s, a, a'}(T^{(i)}) \|$
            \STATE $m \gets m - 2(m \cdot \hat{n}) \hat{n}$
        \ENDIF
    \ENDFOR
    \STATE $w \gets w + \alpha m$ 
    \STATE $m \gets m - \frac{1}{2}\alpha \nabla E(w)$ \hfill \textit{// Final momentum half step}
\end{algorithmic}
\end{algorithm}

A final clean-up step ensures that if the leapfrog integration terminates outside the feasible region, the sample is immediately rejected. This guarantees that every accepted sample satisfies the $\epsilon$-ball property for each state $s \in \mathcal{D}$. To ensure initial feasibility, we warm-start the HMC chain with the point estimate $\widehat{T}^*$ obtained via ITL in Algorithm \ref{algo_ITL} (See below). By reflecting off boundaries rather than blindly rejecting, our algorithm achieves a rejection rate of about 20\% to 60\% across our experiments—a significant improvement over the nearly 100\% rejection rate of standard rejection sampling.

\begin{algorithm}[H]
    \caption{Point estimate $\widehat{T}^*$ (ITL)}
    \label{algo_ITL}
    \begin{algorithmic}[1] 
        \STATE $i \gets 0$
        \STATE $\pi^{(0)}(\cdot|s) \gets \widehat{\pi}_{\epsilon}(\cdot|s; T^*)$ for $s \in \mathcal{D}$
        \STATE $\pi^{(0)}(\cdot|s) \gets \text{uniform distribution}$ for $s \notin \mathcal{D}$
        \STATE $T^{(0)} \gets T^{\text{MLE}}$
        \STATE $T^{(1)} \gets \textit{solve}_{ITL}(\pi^{(0)}, T^{(0)})$
        \STATE $i \gets i + 1$
        \WHILE{$T^{(i)}$ does not satisfy $\epsilon$-ball property for each $s \in \mathcal{D}$}
            \STATE $\pi^{(i)}(\cdot|s) \gets \widehat{\pi}_{\epsilon}(\cdot|s; T^*)$ for $s \in \mathcal{D}$
            \STATE $\pi^{(i)}(\cdot|s) \gets \pi^*(T^{(i)})$ for $s \notin \mathcal{D}$
            \STATE $T^{(i + 1)} \gets \textit{solve}_{ITL}(\{\pi^{(k)}, T^{(k)}\}_{k=0}^{i})$
            \STATE $i \gets i + 1$
        \ENDWHILE
    \end{algorithmic}
\end{algorithm}

\textbf{Inverse Transition Learning} 
In addition to our posterior estimate, we can also infer a point estimate $\widehat{T}^*$. By introducing a quadratic loss that enforces such an estimate to remain close to $T^{\text{MLE}}$, and by linearizing \textbf{Constraints 1 \& 2} using a fixed reference policy $\tilde{\pi}$ and fixed transition dynamics $\tilde{T}$, we form a quadratic convex optimization problem that can be solved efficiently using CVXPY \citep{diamond2016cvxpy}:

\begin{align}
    &\min_{T}\sum_{(s, a, s')} N_{s, a, s'} \cdot \left[T(s' \mid s, a) - T^{\text{MLE}}(s' \mid s, a)\right]^2 \nonumber \\
    &\text{subject to} \quad \forall (s,a) \in \mathcal{D}, \forall a' \notin \hat{\epsilon}(s; T^*): \nonumber \\
    &\quad R(s, a) - R(s, a') + \gamma \left(T(\cdot \mid s, a) - T(\cdot \mid s, a')\right)^\top \nonumber \\
    &\quad \times \left(I - \gamma \tilde{T}_{\tilde{\pi}}\right)^{-1} R_{\tilde{\pi}} \geq \epsilon, \nonumber \\
    &\text{and subject to} \quad \forall (s,a) \in \mathcal{D}, \forall a' \in \hat{\epsilon}(s; T^*): \nonumber \\
    &\quad \left| R(s, a) - R(s, a') + \gamma \left(T(\cdot \mid s, a) - T(\cdot \mid s, a')\right)^\top \right. \nonumber \\
    &\quad \left. \times \left(I - \gamma \tilde{T}_{\tilde{\pi}}\right)^{-1} R_{\tilde{\pi}} \right| \leq \epsilon
    \label{eq:point_estimate}
\end{align}

In general, we denote the result of this optimization framework as \( \textit{solve}_{ITL}(\tilde{\pi}, \tilde{T}) \). Because the variables $\tilde{\pi}$ and $\tilde{T}$ are held constant, the problem remains strictly convex with respect to the optimization variable $T$. For the first iteration, we initialize these fixed inputs as $\tilde{\pi} = \widehat{\pi}_{\epsilon}$ and $\tilde{T} = T^{\text{MLE}}$ because they provide the most reliable empirical estimates for the state-action pairs $(s, a)$ where we have data (i.e., $(s, a) \in \mathcal{D}$). This convex formulation avoids the alternative local optima associated with gradient-based methods like MCE \citep{herman2016inverse} and significantly accelerates training.

This yields the iterative procedure described in Algorithm \ref{algo_ITL}. We initialize our constraints using $\widehat{\pi}_{\epsilon}$ and \( T^{\text{MLE}} \). In subsequent steps, \( \textit{solve}_{ITL}(\{\pi^{(k)}, T^{(k)}\}_{k=0}^{i}) \) denotes solving the optimization problem with additional constraints accumulated from all iterations up to $i$. We keep augmenting the constraint set until we reach a point estimate $\widehat{T}^*$ that satisfies the $\epsilon$-ball property for each state $s \in \mathcal{D}$ (Definition \ref{eps ball property}). In essence, $\widehat{T}^*$ represents the dynamics that best fit the batch data $\mathcal{D}$ while also explaining the $\epsilon$-optimal expert behavior.

\section{EXPERIMENTAL SETUP}
\textbf{Environments}
Our evaluations span 11 synthetic environments and a real-life ICU setting. Specifically, we use a 25 states Gridworld with four actions, 10 different 15-states Randomworld with five actions each, and a healthcare scenario focusing on ICU patients with hypotension, utilizing the MIMIC-IV dataset \citep{johnson2020mimic}. In Gridworld, we generate 100 batches of data $\mathcal{D}$, each consisting of five episodes with 15 steps each. For each 10 of the Randomworld, 50 batches are generated, each containing three episodes of ten steps, due to their smaller size. We also evaluate our methods across varying "Coverage \%"  levels in these environments, defined as the percentage of states observed within each batch data $\mathcal{D}$ and various $\epsilon$ values to see how our method compare to baselines for various degrees of sub-optimality. Detailed descriptions of both the synthetic and real-world environments are available in Appendix \ref{environment} and \ref{ICUenvironment} respectively.

In synthetic scenarios, we simulate a suboptimal expert, where approximately $40\%$ of states are stochastic-policy states (See definition \ref{policy states}), demonstrating the robustness of our approach to suboptimal expert behavior. We also include in the Appendix (Table \ref{tab:combined_results_20}, \ref{tab:combined_results_0}, Figure \ref{fig:combined_plots_20} and \ref{fig:combined_plots_0}), results for $20\%$ and $0\%$ (fully optimal expert) of stochastic-policy states which correspond to lower $\epsilon$ values. Note that increased optimality from the expert (lower $\epsilon$ values) leads to less stochasticity (ie. less  stochastic-policy states) amongst the actions selected in the batch data $\mathcal{D}$. Our method does not require the expert to be perfectly optimal and can adapt to various degrees of optimality through $\epsilon$. In the results below, we use $\gamma = 0.95$ and smoothing parameter defined in Equation \ref{MLE baseline} $\delta = 0.001$. 

\textbf{Baselines.} Our methods, Bayesian Inverse Transition Learning (BITL) and Inverse Transition Learning (ITL), are evaluated against \(T^{MLE}\) (Maximum Likelihood Estimation, MLE), \(T^{MCE}\) (Maximum Causal Entropy, MCE) \citep{herman2016inverse}, and a non-expert-informed posterior \(P(T^*|\mathcal{D})\) (PS).

\textbf{Metrics.} We use the following metrics to evaluate learnt dynamics $T's$ and induced policies \(\pi's\):

\textbf{Best/\(\epsilon\)-ball action matching:} 
Proportion of states where the best or \(\epsilon\)-good action is taken:
\begin{align*}
\text{Best: } & \frac{1}{|\mathcal{S}|} 
    \sum_{s \in \mathcal{S}} 
    \mathcal{I}\!\left\{ 
        \arg\max_{a} \pi(a|s) = \arg\max_{a} \pi^*(a|s) 
    \right\}, \\[6pt]
\epsilon\text{-ball: } & \frac{1}{|\mathcal{S}|} 
    \sum_{s \in \mathcal{S}} 
    \mathcal{I}\!\left\{ 
        \arg\max_{a} \pi(a|s) \in \epsilon(s; T^*) 
    \right\}.
\end{align*}

\textbf{Bayesian regret:} 
     Regret of a sample-based empirical posterior distribution $\widehat{P}(T|\mathcal{D})$, where $|\widehat{P}(T|\mathcal{D})|$ is number of T's in the sample-based distribution:
    \begin{align*}
& \frac{1}{|\widehat{P}(T|\mathcal{D})|^2} 
   \sum_{T \in \widehat{P}(T|\mathcal{D})} 
   \sum_{T' \in \widehat{P}(T|\mathcal{D})} \\
& \quad \times 
   \mathbb{E}_{s_0 \sim \mu_0}\!\left[
      \big| V^{\pi^*(T)}(s_0; T) 
        - V^{\pi^*(T')}(s_0; T) \big|
   \right]
\end{align*}

We also report the (normalized) \textbf{Value}, calculated as the value achieved by the optimal policy of the learned dynamics under the true dynamics $T^*$, i.e., $\mathbb{E}_{s_0}[V^{\pi^*(T)}(s_0; T^*)]$. In addition, we report the \textbf{CVaR} of Value across datasets, \textbf{Total Variation}, the \textbf{number of violated constraints (Inequalities~\ref{eq: constraints1}, \ref{eq: constraints2})}, and training \textbf{Time} (in seconds) for both the baseline MCE and our ITL method. Finally, \textbf{Value} and \textbf{Best/$\epsilon$-ball action matching} are evaluated in both standard and transfer settings.

\vspace{-0.1cm}
\paragraph{Transfer Task Setting}
In this paper, a transfer task refers to the evaluation of the learned dynamics and/or policy under a reward function different from the one used during training. This setting is particularly relevant in healthcare, where the transition dynamics, which describe the physiological response of the human body to treatments, remain largely consistent. However, the reward function, which captures the clinical objectives being optimized, can vary across hospitals, patient populations, and individual clinicians. By evaluating our method under different reward functions, we assess its robustness to variations in clinical goals, ensuring that the learned dynamics remain useful even when the optimization criteria shift. This aligns with real-world applications, where treatment strategies may change while the underlying patient physiology remains unchanged. To demonstrate that our learned dynamics can adapt to actions not seen in the demonstrations, we set up a transfer task where the optimal actions under this new reward function differ from those in the original task (See Figure \ref{fig:gridworld}, \ref{fig:gridworldtransfer}). This ensures the new task requires actions we didn't observe in the initial demonstrations. This tests the model's adaptability to changes in task conditions.

\textbf{Computing the results.} For each dynamics \(T\), we compute its optimal policy \(\pi^*(T)\) using Value Iteration \citep{sutton2018reinforcement}, when needed to compute metrics in terms of policies. For posterior distributions \(P(T|\mathcal{D})\) and \(P_{\epsilon}(T|\mathcal{D})\), we determine the optimal policies of 5,000 sampled \(T's\) and average them.

\textbf{How to Tune \(\epsilon\):} In the healthcare setting, we only have access to an offline dataset. To select an appropriate \(\epsilon\), we evaluate the performance of ITL on a held-out validation set. We choose the \(\epsilon\) that performs best on this validation set based on the \textbf{Best/\(\epsilon\)-ball action matching metric} for both standard and transfer tasks (defined by different reward functions). As shown in Figure \ref{fig:epsilon tuning} (See Appendix), \(\epsilon = 5\) appears to be suitable, and this value is used for training and computing the results in the real-life healthcare experiments (See Table \ref{tab:method_comparison_hypo_5}). We also present results for \(\epsilon = 10\) and \(\epsilon = 15\), as these values are also reasonable. Results for \(\epsilon = 10\) and \(\epsilon = 15\) are provided in Tables \ref{tab:method_comparison_hypo_10} and \ref{tab:method_comparison_hypo_15} in the Appendix, showing similar performance.

\begin{figure*}[htb!]
    \centering
    \begin{minipage}[b]{1.0\textwidth}
        \centering
\includegraphics[width=\textwidth]{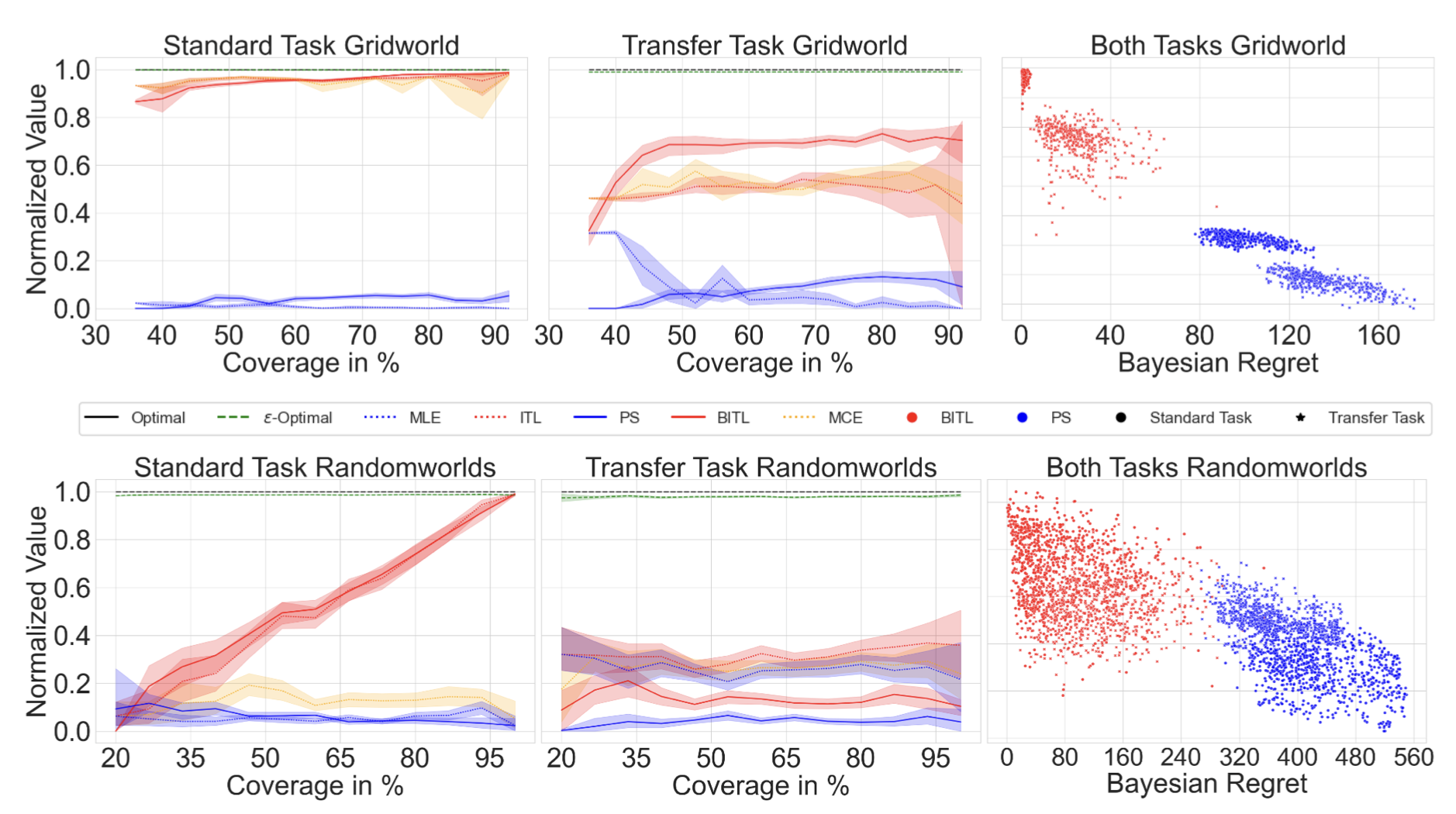}
    \end{minipage}
    \caption{Top row: Normalized Value vs. Coverage for Gridworld (left: Standard Task, middle: Transfer Task), Bottom row: Normalized Value vs. Coverage for Randomworlds (left: Standard Task, middle: Transfer Task). Rightmost plots: Normalized Value vs. Bayesian Regret of both Tasks (top: Gridworld, bottom: Randomworlds). Overall ITL and BITL outerperforms other baselines and BITL help predicting when we can transfer well.}
    \label{fig:combined_plots}
\end{figure*}

\begin{table*}[htb!] 
    \centering
    \small 
    \setlength{\tabcolsep}{4pt} 
    \begin{tabular}{lrrrrr}
        \toprule
        \multicolumn{6}{c}{\textit{Gridworld}} \\
        \cmidrule(lr){1-6}
        Method & \textbf{MCE} & \textbf{BITL} & \textbf{ITL} & \textbf{MLE} & \textbf{PS} \\
        \midrule
        $\epsilon$ matching & $0.65 \pm 0.12$ & $\mathbf{0.78 \pm 0.08}$ & $0.65 \pm 0.12$ & $0.37 \pm 0.06$ & $0.38 \pm 0.02$ \\
        $\epsilon$ matching transfer & $0.5 \pm 0.11$ & $\mathbf{0.64 \pm 0.08}$ & $0.51 \pm 0.11$ & $0.39 \pm 0.07$ & $0.39 \pm 0.02$ \\
        Best matching & $0.55 \pm 0.11$ & $\mathbf{0.64 \pm 0.08}$ & $0.56 \pm 0.11$ & $0.31 \pm 0.06$ & $0.29 \pm 0.02$ \\
        Best matching transfer & $0.45 \pm 0.1$ & $\mathbf{0.54 \pm 0.08}$ & $0.45 \pm 0.11$ & $0.34 \pm 0.07$ & $0.31 \pm 0.02$ \\
        Time & $117.47 \pm 26.34$ & $14384.26 \pm 489.49$ & $0.32 \pm 0.26$ & $\mathbf{0.01 \pm 0.00}$ & $8.65 \pm 0.24$ \\
        Total Variation & $137.62 \pm 4.39$ & $159.36 \pm 3.98$ & $\mathbf{137.05 \pm 4.32}$ & $141.37 \pm 3.8$ & $160.05 \pm 4.43$ \\
        Value CVaR 5\% & $108.76 \pm 26.54$ & $109.04 \pm 4.26$ & $\mathbf{109.7 \pm 9.01}$ & $-4.71 \pm 0.47$ & $-2.35 \pm 1.35$ \\
        Nbr constraints violated & $3.06 \pm 3.68$ & $\mathbf{0.0 \pm 0.0}$ & $\mathbf{0.0 \pm 0.0}$ & $23.13 \pm 6.59$ & $16.37 \pm 3.21$ \\
        \toprule
        \multicolumn{6}{c}{\textit{Randomworlds}} \\
        \cmidrule(lr){1-6}
        Method & \textbf{MCE} & \textbf{BITL} & \textbf{ITL} & \textbf{MLE} & \textbf{PS} \\
        \midrule
        $\epsilon$ matching & $0.54 \pm 0.12$ & $0.75 \pm 0.12$ & $\mathbf{0.76 \pm 0.13}$ & $0.43 \pm 0.11$ & $0.3 \pm 0.05$ \\
        $\epsilon$ matching transfer & $0.46 \pm 0.12$ & $0.38 \pm 0.1$ & $\mathbf{0.5 \pm 0.13}$ & $0.46 \pm 0.12$ & $0.31 \pm 0.05$ \\
        Best matching & $0.38 \pm 0.13$ & $0.57 \pm 0.12$ & $\mathbf{0.58 \pm 0.15}$ & $0.29 \pm 0.11$ & $0.19 \pm 0.05$ \\
        Best matching transfer & $0.34 \pm 0.13$ & $0.26 \pm 0.09$ & $\mathbf{0.37 \pm 0.15}$ & $0.34 \pm 0.13$ & $0.2 \pm 0.05$ \\
        Time & $36.61 \pm 26.36$ & $5561 \pm 223.43$ & $0.65 \pm 0.39$ & $\mathbf{0.0 \pm 0.0}$ & $2.61 \pm 0.36$ \\
        Total Variation & $111.08 \pm 2.42$ & $123.3 \pm 4.06$ & $\mathbf{102.02 \pm 4.84}$ & $111.07 \pm 2.3$ & $127.02 \pm 2.66$ \\
        Value CVaR 5\% & $-459.71 \pm 23.48$ & $-364.34 \pm 34.06$ & $\mathbf{-366.43 \pm 16.03}$ & $-481.98 \pm 23.31$ & $-434.24 \pm 17.22$ \\
        Nbr constraints violated & $11.81 \pm 6.47$ & $\mathbf{0.0 \pm 0.0}$ & $\mathbf{0.0 \pm 0.0}$ & $17.23 \pm 6.75$ & $11.66 \pm 3.12$ \\
        \bottomrule
    \end{tabular}
    \caption{Gridworld and Randomworld results. BITL
performs best in Gridworld, ITL in Randomworld.}
\label{tab:main_synthetic}
\end{table*}

\section{RESULTS}

\textbf{ITL consistently outperforms baseline methods across all metrics and coverage settings.}
Table~\ref{tab:main_synthetic} (synthetic environments), Table~\ref{tab:method_comparison_hypo_5} (ICU dataset), and Figure~\ref{fig:combined_plots} show that ITL and BITL outperform baselines across metrics and coverage levels in both synthetic and real healthcare settings. Additional results for other $\epsilon$ values are provided in the appendix (Tables~\ref{tab:combined_results_20}, \ref{tab:combined_results_0}, Figures~\ref{fig:combined_plots_20}, \ref{fig:combined_plots_0}). In Randomworlds, BITL and ITL clearly surpass MCE, which itself outperforms MLE and PS. In Gridworld, ITL slightly outperforms MCE on average but achieves substantially better worst-case performance, underscoring the value of enforcing hard constraints and the risk of MCE getting stuck in poor local optima due to gradient-based optimization. In the ICU hypotension management task, ITL and BITL significantly outperform all baselines (Table~\ref{tab:method_comparison_hypo_5}), demonstrating the robust applicability of our approach to high-stakes, data-limited clinical settings. ITL also trains substantially faster. In the standard task, ITL and BITL are the only methods that converge to optimal performance as coverage increases (Figure~\ref{fig:combined_plots}, left column), achieving higher values without constraint violations—thanks to our hard-constraint design versus MCE’s soft constraints.

\textbf{ITL enables reliable, fast, and data-efficient dynamics estimation.} 
In Gridworld, MCE required far more training time than ITL (137s vs.~1s) and produced poorer CVaR values, reflecting the difficulties of gradient optimization in non-convex settings. ITL’s efficiency also holds in Randomworld, with training times of 0.7s versus 35s for MCE (Table~\ref{tab:combined_results_40}). Our methods consistently converge to optimal performance as coverage increases, aligning with expert demonstrations. MCE’s tendency to get stuck in local optima further underscores the efficiency and reliability of ITL and BITL in complex environments.

\textbf{Our policies stay within the support of expert actions in \emph{every} visited state, with hard constraints yielding better worst-case outcomes.} 
By enforcing hard constraints, our method ensures policies adhere to expert actions in every visited state, recovering an $\epsilon$-optimal action and minimizing violations. In contrast, MLE, PS, and MCE occasionally breach constraints (Tables~\ref{tab:combined_results_40}, \ref{tab:method_comparison_hypo_5}), while ITL and BITL maintain strict compliance, enhancing reliability. CVaR results (Table~\ref{tab:combined_results_40}) show ITL and BITL achieve stronger worst-case performance by enforcing the $\epsilon$-ball property on estimated dynamics (Definition~\ref{eps ball property}), preserving robustness even in challenging real-world scenarios such as ICU hypotension management.



\textbf{Our method integrates efficiently with Bayesian inference, enabling exploration and predicting transfer success.} 
By inferring calibrated uncertainties, BITL leverages expert knowledge to enhance performance in environments like Gridworld, where it avoids repeated suboptimal decisions (e.g., wall collisions) and achieves the best overall results. This shows that modeling uncertainty over dynamics while incorporating demonstrations provides a clear advantage over point-estimate methods. The issue is less pronounced in Randomworld and the ICU, highlighting the adaptability of our approach across different dynamics. The Bayesian framework also enables more accurate predictions of which tasks will yield successful outcomes based on the learned distributions. Regret plots (Figure~\ref{fig:combined_plots}, right column) show a strong correlation between Bayesian regret values and task performance, with higher regret in transfer tasks than in standard tasks, as expected. Table~\ref{tab:method_comparison_hypo_5} further illustrates this in the ICU dataset, where tasks with lower regret align with better performance.

\begin{figure}[htb!]
    \centering
    \includegraphics[width=\linewidth]{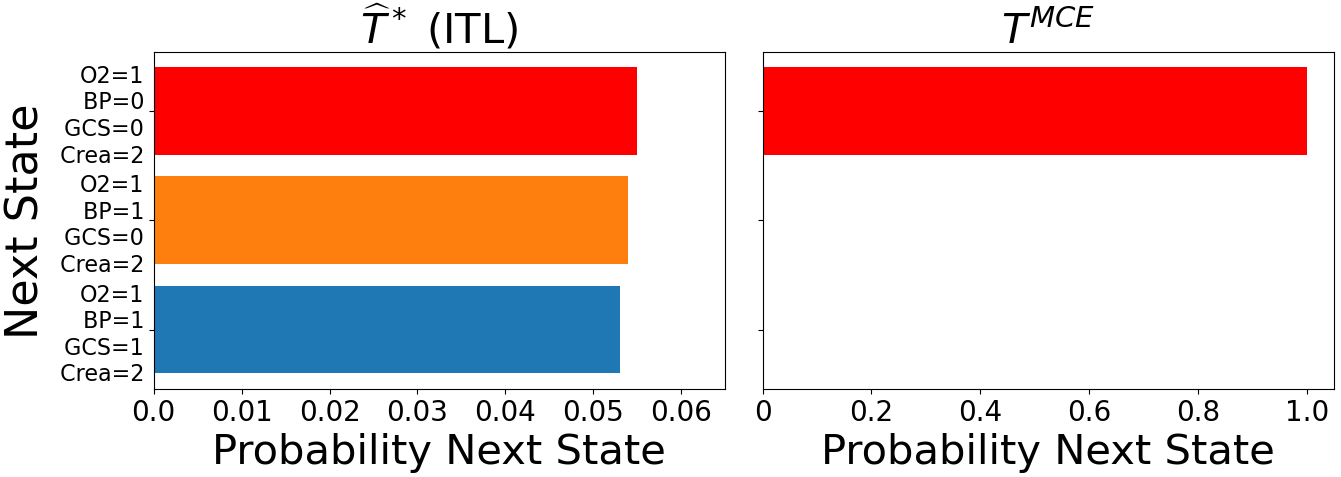}
    \caption{Most likely next three states after IV in state
    O$_2$=1, BP=1, GCS=1, Crea=2. MLE omitted (uniform). ITL spreads mass over improving/maintaining states; MCE collapses onto one state.}
    \label{fig:mimic_counterfactual}
\end{figure}

\textbf{Our method applies to continuous healthcare settings and provides insights through counterfactual reasoning.} 
In the ICU environment, we discretize states using Oxygen ratio (O2), Blood pressure (BP), Creatinine (Crea), and Glasgow Coma Scale (GCS) (Table~\ref{table:discretization_bins} in Appendix). After prescribing IV treatment in state (O2=1, BP=1, GCS=1, Crea=2), we examine the three most likely next states under MLE, ITL, and MCE. Since clinicians never prescribed IV here, MLE gives uniform predictions and is omitted. ITL spreads probability across states that improve or maintain BP and creatinine, while MCE collapses all probability on a single state—unrealistic given physiological complexity and missing data. This illustrates how MCE can yield implausible results in real-world settings. ITL’s next-state probabilities in Figure~\ref{fig:mimic_counterfactual} appear near-uniform due to the prior and absence of data, yet ITL still ranks states and quantifies uncertainty more informatively than MLE (exactly uniform) or MCE (which reduces uncertainty to zero).

\begin{table}[htb!]
    \centering
    \small 
    \setlength{\tabcolsep}{4pt} 
    \begin{tabular}{lrrrrr}
        \toprule
        Method & \textbf{MLE} & \textbf{ITL} & \textbf{MCE} & \textbf{BITL} & \textbf{PS} \\
        \midrule
        \multicolumn{6}{l}{\textbf{\textit{Standard Task}}} \\ 
        Best match.   & 0.33 & \textbf{0.51} & 0.31 & 0.49 & 0.31 \\
        $\epsilon$ match.   & 0.52 & \textbf{1.0}    & 0.56 & \textbf{1.0}    & 0.51 \\
        Constraints & 61   & \textbf{0}    & 47   & \textbf{0}    & 58   \\
        Time (s)        & -    & \textbf{2.23} & 118  & -    & -    \\
        Bayes Regret & -    & -    & -    & \textbf{2.49} & 10   \\
        \midrule
        \multicolumn{6}{l}{\textbf{\textit{Transfer Task}}} \\
        Best match.   & 0.34 & \textbf{0.52} & 0.32 & 0.47 & 0.34 \\
        $\epsilon$ match.   & 0.58 & \textbf{0.97} & 0.68 & 0.90 & 0.58 \\
        Bayes Regret & -    & -    & -    & \textbf{2.80} & 5.40 \\
        \bottomrule
    \end{tabular}
    \caption{ICU results ($\epsilon = 5$). ITL beats all point-estimates and BITL surpasses all Bayesian-estimates.}
    \label{tab:method_comparison_hypo_5}
\end{table}
\section{DISCUSSION}

\textbf{Summary.} 
We addressed the challenge of estimating transition dynamics $T^*$ from near-optimal expert trajectories in offline model-based RL. Our approach, Inverse Transition Learning (ITL), leverages limited expert coverage and near-optimality to estimate $T^*$. We introduced a novel constraint-based method and integrated it within a Bayesian framework to learn a posterior over the dynamics. To our knowledge, this is the first method to combine posterior estimation of dynamics with expert demonstrations. ITL improves decision-making in both synthetic environments and real-world healthcare, such as ICU hypotension management. It not only enhances decision quality but also provides insights into task transfer.

\textbf{Future Work.} 
Although ITL provides a robust approach for estimating dynamics from demonstrations, it is currently limited to discrete, fully observable state spaces. Future work could extend ITL to high-dimensional, continuous, or partially observable settings, along with deeper theoretical analysis.
\subsubsection*{Acknowledgments}
We thank the fantastic
AISTATS reviewers for their constructive comments
and suggestions. This material is based upon work supported by the National Science Foundation under Grant No. IIS-2007076 as well as the Cooperative AI Foundation. Any opinions, findings, and conclusions or recommendations expressed in this material are those of the authors and do not necessarily reflect the views of the National Science Foundation nor the Cooperative AI Foundation.

\appendix
\bibliography{main}
\thispagestyle{empty}

\clearpage
\section*{Checklist}

\begin{enumerate}
    \item For all models and algorithms presented, check if you include:
    \begin{enumerate}
        \item[(a)] A clear description of the mathematical setting, assumptions, algorithm, and/or model. [Yes]
        \item[(b)] An analysis of the properties and complexity (time, space, sample size) of any algorithm. [Not Applicable]
        \item[(c)] (Optional) Anonymized source code, with specification of all dependencies, including external libraries. [No]
    \end{enumerate}

    \item For any theoretical claim, check if you include:
    \begin{enumerate}
        \item[(a)] Statements of the full set of assumptions of all theoretical results. [Not Applicable]
        \item[(b)] Complete proofs of all theoretical results. [Not Applicable]
        \item[(c)] Clear explanations of any assumptions. [Not Applicable]
    \end{enumerate}

    \item For all figures and tables that present empirical results, check if you include:
    \begin{enumerate}
        \item[(a)] The code, data, and instructions needed to reproduce the main experimental results (either in the supplemental material or as a URL). [No]
        \item[(b)] All the training details (e.g., data splits, hyperparameters, how they were chosen). [Yes]
        \item[(c)] A clear definition of the specific measure or statistics and error bars (e.g., with respect to the random seed after running experiments multiple times). [Yes]
        \item[(d)] A description of the computing infrastructure used (e.g., type of GPUs, internal cluster, or cloud provider). [Not Applicable]
    \end{enumerate}

    \item If you are using existing assets (e.g., code, data, models) or curating/releasing new assets, check if you include:
    \begin{enumerate}
        \item[(a)] Citations of the creator if your work uses existing assets. [Yes]
        \item[(b)] The license information of the assets, if applicable. [Not Applicable]
        \item[(c)] New assets either in the supplemental material or as a URL, if applicable. [Not Applicable]
        \item[(d)] Information about consent from data providers/curators. [Not Applicable]
        \item[(e)] Discussion of sensible content if applicable, e.g., personally identifiable information or offensive content. [Not Applicable]
    \end{enumerate}

    \item If you used crowdsourcing or conducted research with human subjects, check if you include:
    \begin{enumerate}
        \item[(a)] The full text of instructions given to participants and screenshots. [Not Applicable]
        \item[(b)] Descriptions of potential participant risks, with links to Institutional Review Board (IRB) approvals if applicable. [Not Applicable]
        \item[(c)] The estimated hourly wage paid to participants and the total amount spent on participant compensation. [Not Applicable]
    \end{enumerate}
\end{enumerate}

\onecolumn
\aistatstitle{Bayesian Inverse Transition Learning:\\ Learning Dynamics from Near-Optimal Trajectories \\
Supplementary Materials}

\section{Non-Convex Feasible Region}
\label{non-convex}
\begin{figure}[h]
    \centering
    \includegraphics[width=0.5\textwidth]{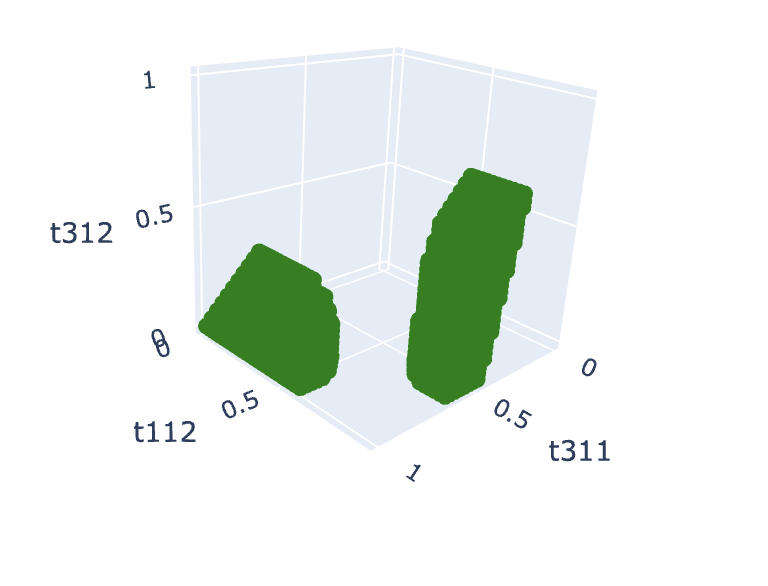}
    \caption{Example of a subspace of the feasible region defined by the constraints on \(T\). The label 'tsas'' indicates the probability of transitioning to state \(s'\) after being in state \(s\) and taking action \(a\).}
    \label{fig: nonconvex}
\end{figure}

When plotting our constraints (equations \ref{eq: constraints1} and \ref{eq: constraints2}) within a three-dimensional subspace of a toy example, as illustrated in Figure \ref{fig: nonconvex}, it becomes apparent that the feasible region can be non-convex, primarily due to the inverse operations within the constraints.

\section{Learning Dynamics and Rewards Jointly}
\label{sec: joint_dynamics_rewards_learning}
A key concern is that assuming a known reward function (beyond simple success/failure signals) is often unreasonable, particularly in settings like robotics. To address this, we extend our ITL algorithm to simultaneously learn the dynamics $T$ and rewards $R$.

This extension follows the exact same framework as our main method. However, to ensure the constraints remain linear with respect to both $R$ and $T$, we adopt an iterative approach. Specifically, for the value function terms $(I - \gamma T_\pi)^{-1} R_\pi$, we fix the matrices $T_\pi$ and vectors $R_\pi$ to their values from the previous iteration (initialized with $T^{\text{MLE}}$ and $R_{\text{prior}}$). This allows us to proceed with the same convex optimization procedure as before.

The new loss function becomes:
\begin{equation}
    \min_{T,R} \sum_{s,a,s'} N_{s,a,s'} \left( T(s' \mid s, a) - T^{\text{MLE}}(s' \mid s, a) \right)^2 + \|R - R_{\text{prior}}\|^2
\end{equation}

Here, $R_{\text{prior}}$ represents a sparse prior: $0$ everywhere except at the goal state, which is assigned a value of $10$. Table \ref{tab:joint_learning} summarizes performance on the Gridworld environment averaged over $20$ different datasets, with $\epsilon = 3$. We compare standard ITL (using the true rewards $R$) against ITL-No-R, which learns the reward jointly with the dynamics. Remarkably, ITL-No-R recovers a reward function that produces dynamics with performance comparable to ITL, despite not having access to the true rewards during training.

\begin{table}[h]
    \centering
    \caption{Comparison of ITL (known rewards) and ITL-No-R (jointly learned dynamics and rewards) on the Gridworld environment ($\epsilon = 3$).}
    \label{tab:joint_learning}
    
    \begin{tabular}{lrr}
        \toprule
        \textbf{Metric} & \textbf{ITL} & \textbf{ITL-No-R} \\
        \midrule
        Value & \textbf{115.70 $\pm$ 2.02} & 115.37 $\pm$ 2.02 \\
        Best matching & \textbf{0.65 $\pm$ 0.11} & 0.64 $\pm$ 0.12 \\
        $\epsilon$-matching & \textbf{0.73 $\pm$ 0.14} & \textbf{0.73 $\pm$ 0.14} \\
        Total-Variation & \textbf{132.46 $\pm$ 6.40} & 132.76 $\pm$ 6.44 \\
        Value Transfer & \textbf{41.79 $\pm$ 8.85} & 41.15 $\pm$ 9.95 \\
        Best matching Transfer & \textbf{0.54 $\pm$ 0.14} & \textbf{0.54 $\pm$ 0.13} \\
        $\epsilon$-matching Transfer & \textbf{0.59 $\pm$ 0.09} & \textbf{0.59 $\pm$ 0.09} \\
        Time & 4.82 $\pm$ 2.92 & \textbf{0.61 $\pm$ 0.60} \\
        \bottomrule
    \end{tabular}
    
\end{table}

\section{Synthetic Environments}
\label{environment}

\subsection{Gridworld Environment}
The Gridworld environment is structured as follows:

\begin{itemize}
    \item \textbf{Grid Size}: The world is a grid consisting of $5 \times 5$ tiles, resulting in a total of 25 distinct states.
    \item \textbf{Actions}: At each state, an agent can choose from four possible actions: move right, move up, move left, or move down.
    \item \textbf{Initial State}: The agent always starts from the bottom left corner of the grid, which is designated as the initial state.
    \item \textbf{Goal State}: The objective for the agent is to reach the goal state, located at the top right corner of the grid.
    \item \textbf{Dynamics}:
    \begin{itemize}
        \item \textbf{Intended Actions}: When the agent selects an action, there is an 80\% chance that it will move deterministically to the intended adjacent state.
        \item \textbf{Slipping}: There is a 20\% chance that the agent will slip, leading to a non-deterministic outcome. In such cases, the agent might end up in any one of the four neighboring tiles (right, left, up, down) of the intended state.
        \item \textbf{Wall Interactions}: If an action would result in the agent moving into a wall (the edge of the grid), the agent remains in its current state. This mechanic ensures that the agent does not leave the confines of the grid.
    \end{itemize}
\end{itemize}

The definition of the reward \(R\) in the grid world environment is structured as follows:

\begin{itemize}
    \item \textbf{Soft-Wall Penalty}: If the agent attempts to move across a tile designated as a \textit{soft-wall}, it incurs a penalty of \(-5\) reward points for each attempt. This mechanic discourages the agent from crossing these specific tiles.
    \item \textbf{Movement Penalty}: For every other tile that the agent moves through, it receives a minor penalty of \(-0.1\) reward points. This encourages the agent to find the shortest possible path to the goal.
    \item \textbf{Goal Reward}: Upon successfully reaching the goal state, the agent is awarded \(+10\) reward points. This substantial reward signifies the completion of the episode and serves as the primary incentive for the agent to navigate the grid efficiently.
\end{itemize}

\begin{figure}[h]
\centering
\textbf{Reward function of Gridworld} \\  
\includegraphics[width=0.5\textwidth]{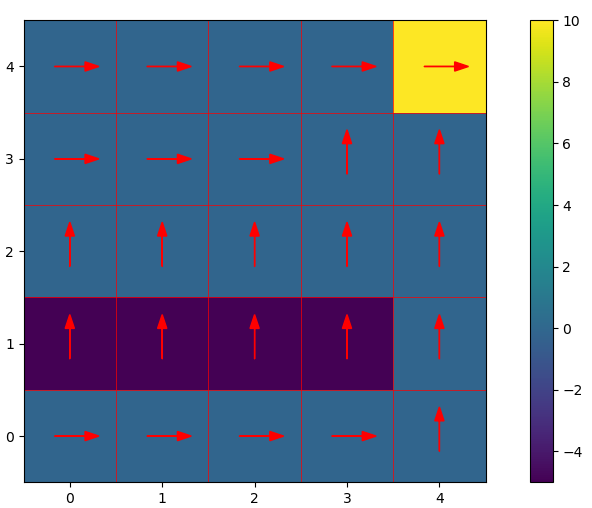}
\caption{Visualization of the grid world environment. Each square in the 5x5 grid represents a unique state, colored based on the associated reward. The 'soft-wall' tiles are distinctively colored to represent a reward of -5. Red arrows on each tile indicate the direction of the optimal policy from that state, leading towards the goal state at the top right corner, which is marked in a different color and has a reward of 10. The starting point is at the bottom left corner, from where the arrows guide the optimal path through the grid.}
\label{fig:gridworld}
\end{figure}

This reward structure is designed to balance the objective of reaching the goal state as quickly as possible with the challenge of navigating around soft-wall tiles. The penalties for unnecessary movements and soft-wall crossings ensure that the agent must carefully consider each action, while the reward for reaching the goal state motivates the agent to complete its objective efficiently. Figure \ref{fig:gridworld} shows the grid world environment, showcasing both the rewards for each state and the optimal policy indicated by red arrows. This environment presents a challenge for an agent to learn the most efficient path from the initial state to the goal state, taking into account the probabilistic nature of movement due to slipping. The deterministic and non-deterministic outcomes necessitate strategic planning and adaptability in the agent's approach to navigating the grid. 

\subsubsection{Transfer Task}

\begin{figure}[h]
\centering
\textbf{Reward function of Gridworld in Transfer Task} \\  
\includegraphics[width=0.5\textwidth]{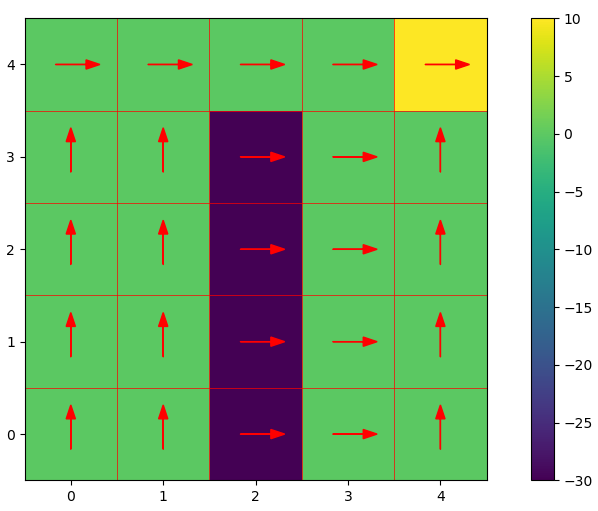}
\caption{The grid world environment after the transfer task modification. The soft-wall tiles, previously located, have been repositioned, visibly changing the reward distribution across the grid. As a result, the optimal policy, indicated by red arrows, now follows a novel route that adapts to the new reward structure, aiming to minimize penalties and maximize returns en route to the goal state at the top right corner.}
\label{fig:gridworldtransfer}
\end{figure}

In the transfer task, we preserve the structure of the original grid world environment, with no alterations to the state space or action set. However, we introduce a significant change to the reward function: the location of the soft-wall is shifted, thereby altering the reward landscape. This modification necessitates the derivation of a new optimal policy that accounts for the updated rewards and navigates the agent from the initial state to the goal state via a different path. See Figure \ref{fig:gridworldtransfer}.

The updated reward function is defined identically to the original environment, with soft-wall tiles incurring a penalty of \(-5\) reward points, other tiles a penalty of \(-0.1\) points, and a reward of \(+10\) points assigned upon reaching the goal state. The shift in the soft-wall location directly affects the agent's trajectory, demonstrating the agent's ability to adjust its policy in response to changes in environmental dynamics. This transfer task effectively evaluates the flexibility and robustness of the learned policy.

\subsection{Randomworld Environment}

We introduce the RandomWorld environment (inspired by \citep{jiang2015dependence}), which is designed to evaluate the performance of reinforcement learning algorithms under conditions of high uncertainty and stochasticity. The RandomWorld environment is characterized by the following properties:

\begin{itemize}
    \item \textbf{State Space}: The environment comprises 15 distinct states.
    \item \textbf{State Space}: The environment comprises 5 distinct actions.
    \item \textbf{Dynamics }: For each state-action pair, 5 successor states are chosen at random to have nonzero transition probability. These probabilities are drawn independently from $\text{Uniform}[0,1]$ and normalized to sum to one. 
    \item \textbf{Initial State Distribution}: The initial state for each episode is selected with uniform probability across all 15 states.
    \item \textbf{Absence of a Goal State}: RandomWorld is devoid of a specified goal state, thus simulating scenarios where an agent's exploration is continuous and without a predetermined endpoint.
    \item \textbf{Reward Function}: Rewards are assigned randomly yet structured such that state 1 yields the highest expected reward, and state 15 the lowest. Specifically, the reward for state \( s \), \( R(s) \), is uniformly distributed within the interval \([16-s -1, 16-s]\), aligning with the descending order of state desirability.
\end{itemize}

The inherent randomness in state transitions and rewards within RandomWorld poses a significant challenge to reinforcement learning strategies, necessitating the development of policies that are robust to uncertainty and variability in environmental dynamics.

\subsubsection{Transfer Task}

In the RandomWorld environment's transfer task, we introduce a  modification to the reward function while preserving all other environmental characteristics. This adjustment is aimed at assessing the adaptability of reinforcement learning algorithms when confronted with a new reward paradigm. The specifics of the transfer task are as follows:

\begin{itemize}
    \item \textbf{Inverted Reward Structure}: We reverse the ranking of state desirability; state 1 is now the least desirable state, and state 15 is the most desirable.
    \item \textbf{Random Reward Generation}: The reward for state \( s \) in the transfer task, \( R_{\text{transfer}}(s) \), is determined by a random draw from a uniform distribution over the range \([s-1, s]\), thus ensuring that higher state numbers correspond to higher expected rewards.
\end{itemize}

This reversal in the reward hierarchy necessitates that the agent recalibrates its policy to align with the new set of rewards. It provides an insightful measure of the algorithm's capacity to adapt to drastic changes in the reward structure within a stochastic environment.

\subsection{Generating batch data $\mathcal{D}$}
\label{batch data}
A critical component of our experiments in offline reinforcement learning is the generation of a batch data $\mathcal{D}$, which is constructed based on a predefined coverage percentage of the state space and a given $\epsilon$, measuring the degree of optimality of the expert $\pi_{\epsilon}(.|. ;T^*)$ with respect to the true unknown dynamics $T^*$. The batch data $\mathcal{D}$ is created through the following procedure:

\begin{enumerate}
    \item \textbf{State Selection}: We randomly select a certain percentage of the total states, corresponding to the coverage parameter, to include in our batch data.
    \item \textbf{Action Selection}: For each state $s$ included in our selection, we identify actions in the $\epsilon$-ball $\epsilon(s;T^*)$, with respect to the $\epsilon$-optimal expert $\pi_{\epsilon}(.|. ;T^*)$.
    \item \textbf{Transition Sampling}: We sample $K$ transitions for each state-action pair $(s, a)$ from the true dynamics $T^*$.
    \item \textbf{Dataset Construction}: The batch dataset $\mathcal{D}$ is comprised of the transitions collected, each represented as a tuple $(s, a, s')$, with $s$ as the state, $a$ as the action, $s'$ as the next state.
\end{enumerate}

In our experimental setup, we define the value of $K$, which dictates the number of transitions sampled for each state-action pair that aligns with the $\epsilon$-optimal expert policy $\pi_{\epsilon}(.|. ;T^*)$. For the Gridworld environment, we set  $K = 10$ , acknowledging the larger state space and the need for a comprehensive dataset that encapsulates the dynamics around the optimal policy. In contrast, for RandomWorld, we set $K = 5$, suitable for its smaller size and complexity. 

\subsection{Averaging Procedure for Experimental Results}

To achieve statistical rigor in our experiments, we average our results over multiple independently generated datasets for both Gridworld and RandomWorld environments:

\begin{itemize}
    \item \textbf{Gridworld}: We generate 100 independent batch datasets for the Gridworld environment. The experimental results for each dataset are recorded, and the final result is obtained by averaging these outcomes. 
    
    \item \textbf{RandomWorld}: For the RandomWorld environment, we create 10 independent instances of the environment. Each of these RandomWorld instances is accompanied by 50 independently generated batch datasets, leading to a total of 500 unique datasets (10 worlds multiplied by 50 datasets each). The experimental outcomes across these datasets are compiled, and their average is computed to determine the overall performance in the RandomWorld environment.
\end{itemize}

This methodology, involving the independent generation of each dataset and each world instance, provides a comprehensive and unbiased evaluation of the algorithms, ensuring that our results are not influenced by any specific configuration or sample of the environment.

\section{Real-life ICU Environment}
\label{ICUenvironment}
Following the experiments within a synthetic environments, we now transition to the evaluation of our methodology in a real-world scenario. To this end, we selected the Medical Information Mart for Intensive Care IV (MIMIC-IV) dataset as our experimental field. This dataset offers a rich, diverse, and challenging setting for testing our method, especially given its potential to contribute to advancements in healthcare analytics and patient care strategies.

\subsection{About MIMIC-IV Dataset}

The MIMIC-IV dataset, developed by the MIT Lab for Computational Physiology and publicly available, aggregates a vast range of anonymized health data from critical care units at Beth Israel Deaconess Medical Center in Boston. Covering over a decade's worth of patient admissions, it provides detailed records on demographics, vital signs, lab tests, medications, and more, establishing itself as a critical resource for healthcare model development. Its comprehensive scope spans all patient care aspects, enabling the creation of holistic models for predicting diverse patient outcomes. The dataset's richness lies in its variety, covering over 40,000 patients of different ages, ethnicities, and conditions, and its granularity, offering high-resolution data points and time-stamped records, which are essential for developing precise, dynamic healthcare models. Moreover, MIMIC-IV's public accessibility fosters a global research community's collaboration, enhancing healthcare analytics advancements.

Utilizing the MIMIC-IV dataset, we showcase out the learning applicability of our method in real-world healthcare, to get valuable insights from the data in such a complicated environemnt.

\subsection{Data Preprocessing for Hypotension Analysis}

In our investigation into hypotension within ICU settings, we tailored our preprocessing steps to exclusively include patients affected by this condition. Our methodology commenced with the application of specific filters on the MIMIC-IV dataset to accurately identify the patient cohort of interest. These filters were designed to capture adults aged 18 to 80 years, who had ICU stays of a minimum duration of 24 hours, and exhibited Mean Arterial Pressure (MAP) readings of 65mmHg or below, indicative of acute hypotension.

The analytical framework of our study is built around a carefully selected set of five clinical variables that constitute the state space, namely: \text{creatinine levels} (\textbf{(Crea)}), \text{Glasgow Coma Scale score} (\textbf{(GCS)}), \text{mean blood pressure} (\textbf{(BP)}), and the ration of \text{partial pressure of oxygen}, over \text{fraction of inspired oxygen} (\textbf{(O2)}). The action space encompasses two primary treatment modalities: \text{intravenous (IV) fluid bolus therapy} and \text{vasopressor therapy}. This precise filtering approach yielded a dataset comprising 1,684 distinct ICU admissions, from which we derived approximately 100,000 tuples $(\text{state, action, next}\_\text{state}) \in \mathcal{D}$. This dataset serves as the foundation to evaluate our method and the baselines.

\subsection{State Space Construction}

The state space for our model is constructed by discretizing five key clinical variables extracted from the MIMIC-IV dataset: partial pressure of oxygen, fraction of inspired oxygen, mean blood pressure, Glasgow Coma Scale (GCS), and creatinine levels. Discretization involves binning these variables into distinct categories based on clinically relevant thresholds, as follows:

\begin{table*}[ht]
\footnotesize
\centering
\caption{Discretization of Clinical Variables into Bins}
\label{table:discretization_bins}
\begin{tabular}{llcc}
\toprule
\textbf{Abbrev} & \textbf{Clinical Variable} & \textbf{Threshold} & \textbf{Bin} \\
& & & \textbf{Value} \\
\midrule
\textbf{O2} & Partial Pressure of Oxygen / Fraction Inspired Oxygen & $\geq 200$ & 0 \\
\textbf{O2} & Partial Pressure of Oxygen / Fraction Inspired Oxygen & $< 200$ and $\geq 100$ & 1 \\
\textbf{O2} & Partial Pressure of Oxygen / Fraction Inspired Oxygen & $< 100$ & 2 \\
\textbf{BP} & Mean Blood Pressure & $\geq 70$ mmHg & 0 \\
\textbf{BP} & Mean Blood Pressure & $< 70$ mmHg & 1 \\
\textbf{GCS} & Glasgow Coma Scale (GCS) & $\leq 12$ & 0 \\
\textbf{GCS} & Glasgow Coma Scale (GCS) & $> 14$ & 1 \\
\textbf{Crea} & Creatinine & $\leq 1.9$ mg/dL & 0 \\
\textbf{Crea} & Creatinine & $> 1.9$ and $\leq 4.9$ mg/dL & 1 \\
\textbf{Crea} & Creatinine & $> 4.9$ mg/dL & 2 \\
\bottomrule
\end{tabular}
\end{table*}

Based on the binning schema presented, the state space comprises all possible combinations of these bins, leading to a total of $3 \times 3 \times 2 \times 2 = 36$ unique states. This structure effectively captures diverse clinical scenarios within a manageable framework for analyzing the dynamics of hypotension treatment. To illustrate the discretization process and the resultant bin mapping, consider a hypothetical patient data point with the following clinical variable values:

\begin{itemize}
    \item Partial Pressure of Oxygen / Fraction Inspired Oxygen: 150
    \item Mean Blood Pressure: 65 mmHg
    \item Glasgow Coma Scale: 10
    \item Creatinine: 5 mg/dL
\end{itemize}

Based on the discretization schema provided in Table \ref{table:discretization_bins}, this patient data point would be mapped to the following bins:

\begin{itemize}
    \item Partial Pressure of Oxygen / Fraction Inspired Oxygen ($150$): Bin 1 (since $100 \leq 150 < 200$)
    \item Mean Blood Pressure ($65$ mmHg): Bin 1 (since $65 < 70$ mmHg)
    \item Glasgow Coma Scale (GCS) ($10$): Bin 0 (since $10 \leq 12$)
    \item Creatinine ($5$ mg/dL): Bin 2 (since $5 \geq 4.9$)
\end{itemize}

Thus, the tuple $(150, 65, 10, 5)$ would be mapped to the discretized state $(1, 1, 0, 2)$ according to our binning process. This discretization approach allows us to capture a comprehensive yet manageable representation of the patient's clinical status, facilitating the application of our offline reinforcement learning model to infer the unknown dynamics $T^*$.

\subsection{Action Space Definition}

The action space in our model encapsulates the range of possible treatments administered to patients suffering from hypotension. It consists of four discrete actions, each representing a specific treatment strategy. The actions are enumerated as follows:

\begin{table*}[ht]
\centering
\begin{tabular}{cl}
\toprule
\textbf{Action} & \textbf{Description} \\
\midrule
0 & No treatment administered \\
1 & Vasopressor therapy administered \\
2 & Intravenous (IV) fluid bolus administered \\
3 & Both vasopressor therapy and IV fluid bolus  \\
\bottomrule
\end{tabular}
\caption{Definition of Actions in the Treatment Strategy Space}
\label{table:action_space}
\end{table*}

Each action is designed to reflect the clinical decisions made in the intensive care unit for managing patients' blood pressure levels. Action 0 (no treatment) represents a conservative approach, where no immediate intervention is applied. Action 1 (vasopressor therapy) and Action 2 (IV fluid bolus) correspond to the administration of specific treatments aimed at increasing blood pressure. 

\subsection{Reward Function Definition}

The reward function, \(R(s)\), quantifies the desirability of each state \(x = (s_1, s_2, s_3, s_4)\) based on the bin values corresponding to the discretized clinical variables (each $s_i$ correspond to a bin number). Formally, the reward function is defined as:
\[ R(s) = 60 - 10 \times (s_1 + s_2 + s_4) \]

This formulation encapsulates our intuition that higher bin values for any of the clinical variables signify a deterioration in the patient's condition, indicating more severe or dangerous vital signs. Consequently, the reward decreases linearly by a factor of 10 for each increment in the bin values of the state components. The choice of this linear penalty ensures a straightforward interpretation of the state's severity, with a base reward of 60 being adjusted downward based on the sum of the bin values in the state.

While this reward function offers a reasonable approximation for assessing the clinical states in the context of hypotension, it is important to acknowledge that other formulations could be equally valid. The essential criterion for any chosen reward function is its ability to accurately differentiate between clinically favorable and unfavorable states, thereby guiding the reinforcement learning model towards optimizing treatment strategies that mitigate the risks associated with hypotension.

\subsection{Transfer Task and Modified Reward Function}

We introduce a transfer task to evaluate the model's adaptability and performance under a different reward function and keeping evrything else identical. The modified reward function for the transfer task is defined as:
\[ R_{\text{transfer}}(s) = 60 - 10 \times ( s_2 + s_4) \]

This adjustment means that the reward now decreases quadratically, rather than linearly, with the values of the features within each state \(s = (s_1, s_2, s_3, s_4)\). The quadratic penalty intensifies the impact of higher bin values, more aggressively penalizing states indicative of worsening patient conditions. This change aims to test the method's sensitivity and response to more severe deteriorations in the clinical variables, pushing the reinforcement learning algorithm to prioritize avoiding high-risk states with even greater emphasis. Such a modification in the reward function's structure is pivotal for assessing the robustness and flexibility of our method. It allows us to explore how different reward formulations can influence decision-making strategies in the context of medical treatment optimization, particularly under scenarios with escalating risks.

\subsection{How to Tune \(\epsilon\):}
\begin{figure*}[t]    \vspace{-1\baselineskip}
    \centering
    \begin{minipage}[b]{0.45\textwidth}
        \centering
        \includegraphics[width=\textwidth]{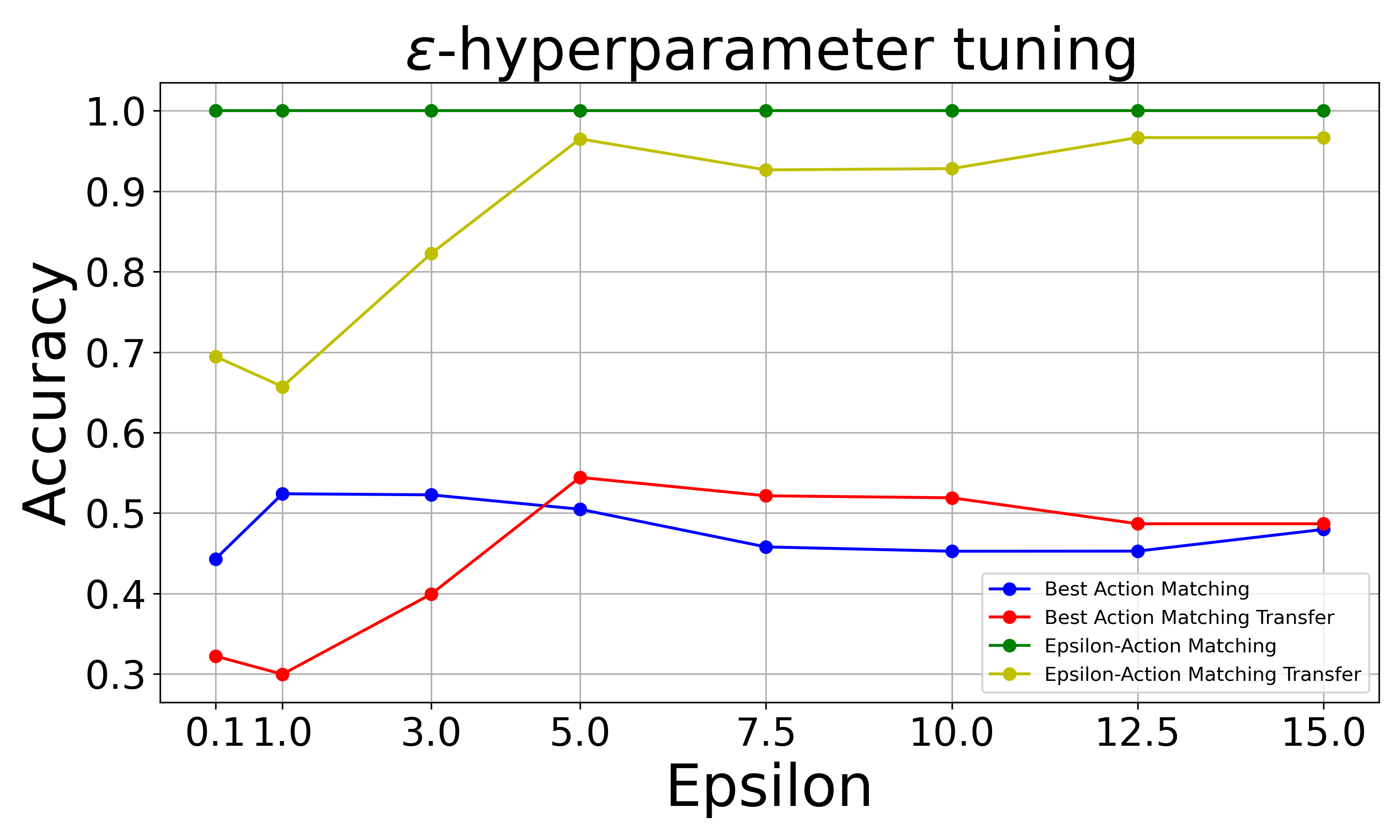}
    \end{minipage}
    \caption{ITL performance across $\epsilon$ values on a held-out dataset for the healthcare setting.}
    \label{fig:epsilon tuning}
\end{figure*}

In the healthcare setting, we only have access to an offline dataset. To select an appropriate \(\epsilon\), we evaluate the performance of ITL on a held-out validation set. We choose the \(\epsilon\) that performs best on this validation set based on the \textbf{Best/\(\epsilon\)-ball action matching metric} for both standard and transfer tasks (defined by different reward functions). As shown in Figure \ref{fig:epsilon tuning}, \(\epsilon = 5\) appears to be suitable, and this value is used for training and computing the results in the real-life healthcare experiments (See Table \ref{tab:method_comparison_hypo_5}). We also present results for \(\epsilon = 10\) and \(\epsilon = 15\), as these values are also reasonable. Results for \(\epsilon = 10\) and \(\epsilon = 15\) are provided in Tables \ref{tab:method_comparison_hypo_10} and \ref{tab:method_comparison_hypo_15} in the Appendix, showing similar performance.

 \subsection{Construction of the \(\epsilon\)-Optimal Expert Policy \(\pi_{\epsilon}\)}

In the absence of explicit knowledge about the true dynamics \(T^*\) governing the environment, our methodology for obtaining an estimate of the \(\epsilon\)-optimal expert policy, \(\widehat{\pi}_{\epsilon}(\cdot|\cdot; T^*)\), leverages the historical batch data \(\mathcal{D}\) collected from the ICU. We define an action \(a\) to be valid for a state \(s\) if and only if action \(a\) was executed in at least 5\% of the instances where state \(s\) was observed in \(\mathcal{D}\). Actions not meeting this criterion are considered invalid for the state, reflecting an approach that filters actions based on their historical prevalence and relevance to specific states.

Leveraging domain knowledge within the critical care domain, we set the \(\epsilon\) parameter to 5 for our experiments. This parameter choice reflects a balance, aiming to capture the degree of optimality in the actions taken by medical professionals in the ICU, under the assumption that the most frequently taken actions represent a near-optimal strategy given the complex dynamics and uncertainties inherent in patient care. Our experimental results demonstrate non-significant changes across various \(\epsilon\) values, suggesting that the selected \(\epsilon\)-optimal policy robustly encapsulates the expert behavior within the dataset, without significant sensitivity to the exact \(\epsilon\) threshold.

\section{Results}

We include in this section the full set of results for the synthetic worlds where we include expert with various degree of optimality leading to 20\% and 0\% of stochastic-policy states) as well as multiple $\epsilon$ values as well for the results with the real healthcare dataset.

\begin{figure*}[ht]
    \centering
    \begin{minipage}[b]{0.9\textwidth}
        \centering
\includegraphics[width=\textwidth]{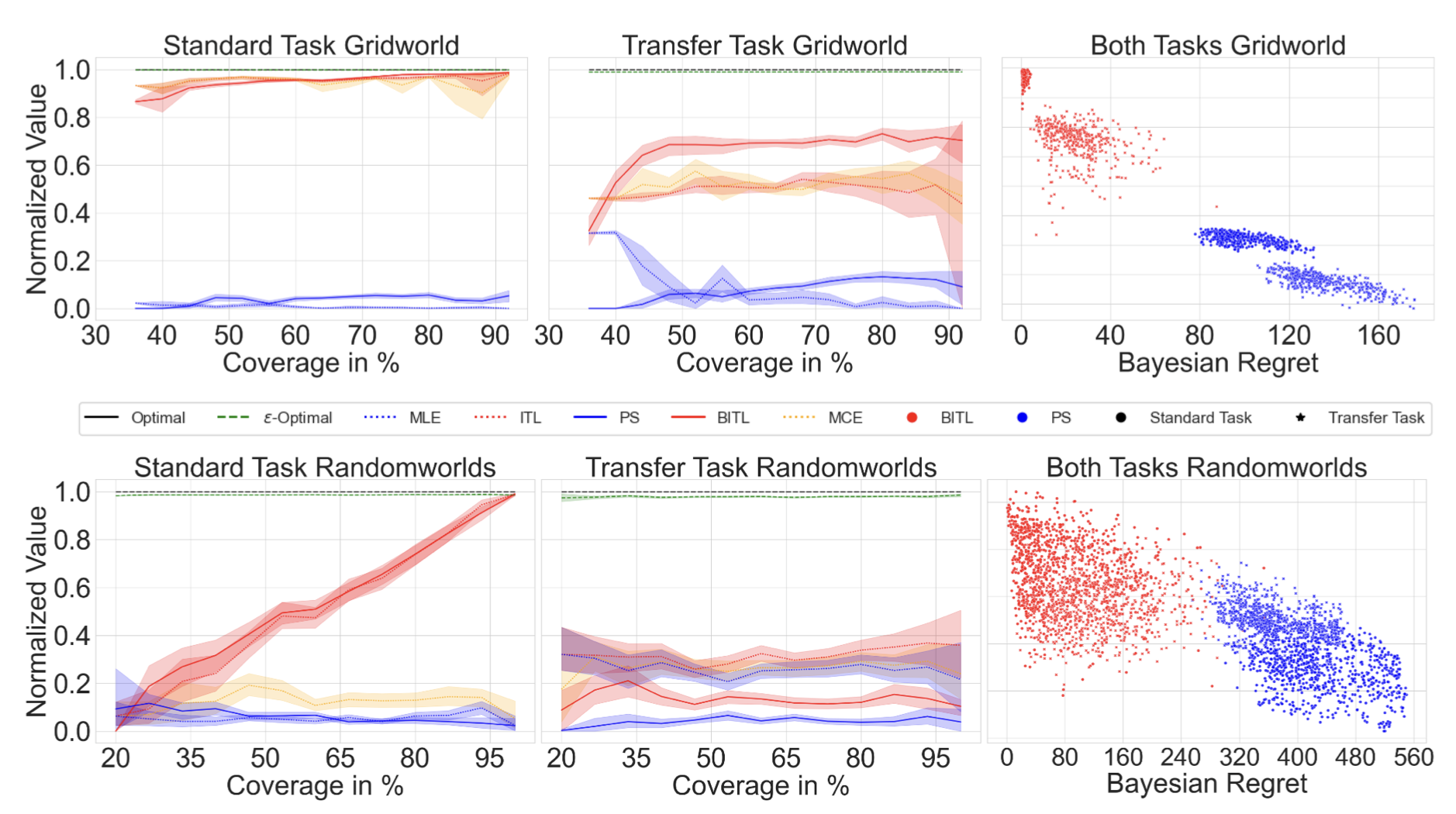}
    \end{minipage}
    \caption{($40\%$ stochastic-policy states, $\epsilon$ Gridworld = 3.0, $\epsilon$'s RandomWorlds = [2.37, 2.53, 2.21, 1.89, 4.27, 1.74, 2.37, 2.84, 2.37, 2.53]) Top row: Normalized Value vs. Coverage for Gridworld (left: Standard Task, middle: Transfer Task), Bottom row: Normalized Value vs. Coverage for Randomworlds (left: Standard Task, middle: Transfer Task). Rightmost plots: Normalized Value vs. Bayesian Regret of both Tasks (top: Gridworld, bottom: Randomworlds).}
    \label{fig:combined_plots_appdx}
\end{figure*}
\begin{figure*}[ht]
    \centering
    \begin{minipage}{\textwidth}
        \centering
        \captionof{table}{Gridworld and Randomworlds Results ($40\%$ stochastic-policy states, $\epsilon$ Gridworld = 3.0, $\epsilon$s RandomWorlds = [2.37, 2.53, 2.21, 1.89, 4.27, 1.74, 2.37, 2.84, 2.37, 2.53])}
        \label{tab:combined_results_40} 
        \resizebox{0.9\textwidth}{!}{ 
        \begin{tabular}{l*{5}{r}}
        \toprule
        \multicolumn{1}{c}{} & \multicolumn{5}{c}{Gridworld} \\
        \cmidrule(lr){2-6}
        Method & MCE & BITL & ITL & MLE & PS \\
        \midrule
        $\epsilon$ matching   & 0.65 ± 0.12 & \textbf{0.78 ± 0.08} & 0.65 ± 0.12 & 0.37 ± 0.06 & 0.38 ± 0.02 \\
        $\epsilon$ matching transfer & 0.5 ± 0.11 & \textbf{0.64 ± 0.08} & 0.51 ± 0.11 & 0.39 ± 0.07 & 0.39 ± 0.02 \\
        Best matching & 0.55 ± 0.11 & \textbf{0.64 ± 0.08} & 0.56 ± 0.11 & 0.31 ± 0.06 & 0.29 ± 0.02 \\
        Best matching transfer & 0.45 ± 0.1 & \textbf{0.54 ± 0.08} & 0.45 ± 0.11 & 0.34 ± 0.07 & 0.31 ± 0.02 \\
        Time & 117.47 ± 26.34 & 14384.26 ± 489.49 & 0.32 ± 0.26 & \textbf{0.01 ± 0.00}  & 8.65 ± 0.24 \\
        Total Variation & 137.62 ± 4.39 & 159.36 ± 3.98 & \textbf{137.05 ± 4.32} & 141.37 ± 3.8 & 160.05 ± 4.43 \\
        Value CVaR 1\% & 56.05 ± 15.08 & \textbf{104.64 ± 5.43} & 103.87 ± 15.67 & -5.34 ± 0.54 & -4.33 ± 1.68 \\
        Value CVaR 2\% & 76.7 ± 17.51 & \textbf{107.36 ± 5.19} & 106.46 ± 12.51 & -5.18 ± 0.51 & -3.49 ± 1.56 \\
        Value CVaR 5\% & 108.76 ± 26.54 & 109.04 ± 4.26 & \textbf{109.7 ± 9.01} & -4.71 ± 0.47 & -2.35 ± 1.35 \\
        Nbr constraints violated & 3.06 ± 3.68 & \textbf{0.0 ± 0.0} & \textbf{0.0 ± 0.0} & 23.13 ± 6.59 & 16.37 ± 3.21 \\
        \midrule
        \multicolumn{1}{c}{} & \multicolumn{5}{c}{Randomworlds} \\
        \cmidrule(lr){2-6}
        Method & MCE & BITL & ITL & MLE & PS \\
        \midrule
        $\epsilon$ matching & 0.54 ± 0.12 & 0.75 ± 0.12 & \textbf{0.76 ± 0.13} & 0.43 ± 0.11 & 0.3 ± 0.05 \\
        $\epsilon$ matching transfer & 0.46 ± 0.12 & 0.38 ± 0.1 & \textbf{0.5 ± 0.13} & 0.46 ± 0.12 & 0.31 ± 0.05 \\
        Best matching & 0.38 ± 0.13 & 0.57 ± 0.12 & \textbf{0.58 ± 0.15} & 0.29 ± 0.11 & 0.19 ± 0.05 \\
        Best matching transfer & 0.34 ± 0.13 & 0.26 ± 0.09 & \textbf{0.37 ± 0.15} & 0.34 ± 0.13 & 0.2 ± 0.05 \\
        Time & 36.61 ± 26.36 & 5561 ± 223.43 & 0.65 ± 0.39 & \textbf{0.0 ± 0.0} & 2.61 ± 0.36 \\
        Total Variation & 111.08 ± 2.42 & 123.3 ± 4.06 & \textbf{102.02 ± 4.84} & 111.07 ± 2.3 & 127.02 ± 2.66 \\
        Value CVaR 1\% & -522.35 ± 5.49 & -423.21 ± 29.84 & \textbf{-404.02 ± 0.13} & -525.94 ± 3.86 & -452.19 ± 22.88 \\
        Value CVaR 2\% & -514.69 ± 7.06 & -398.25 ± 30.3 & \textbf{-397.26 ± 5.34} & -519.63 ± 6.42 & -444.69 ± 20.16 \\
        Value CVaR 5\% & -459.71 ± 23.48 & -364.34 ± 34.06 &\textbf{ -366.43 ± 16.03} & -481.98 ± 23.31 & -434.24 ± 17.22 \\
        Nbr constraints violated & 11.81 ± 6.47 & \textbf{0.0 ± 0.0} & \textbf{0.0 ± 0.0} & 17.23 ± 6.75 & 11.66 ± 3.12 \\
        \bottomrule
        \end{tabular} 
        }
    \end{minipage}
\end{figure*}

\begin{figure*}[ht]
    \centering
    \begin{minipage}[b]{0.9\textwidth}
        \centering
\includegraphics[width=\textwidth]{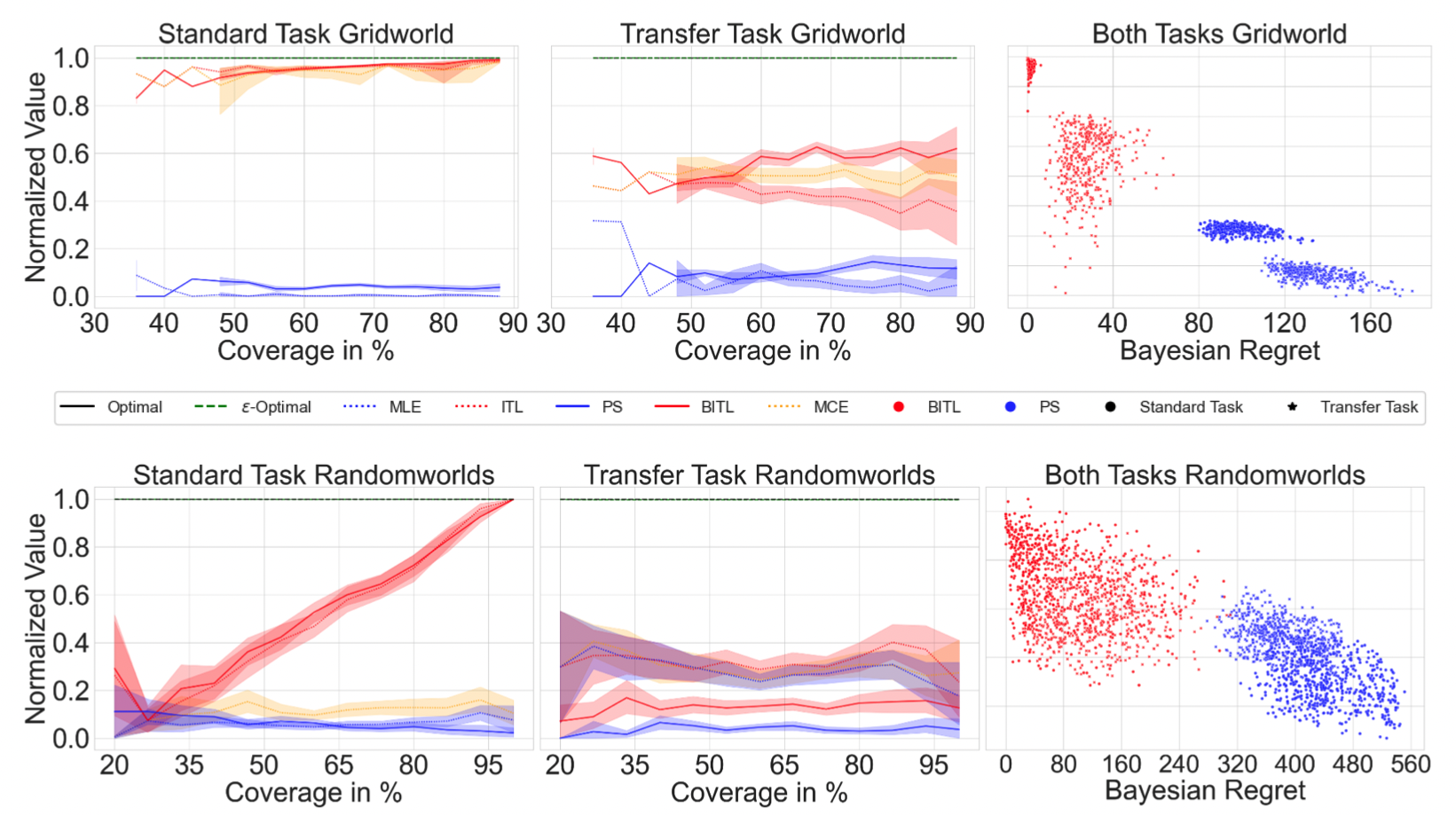}
    \end{minipage}
    \caption{($20\%$ stochastic-policy states, $\epsilon$ Gridworld = 0.3, $\epsilon$s RandomWorlds = [0.79, 1.11, 0.32, 0.63, 1.16, 0.33, 0.57, 0.47, 0.99, 0.32]) Top row: Normalized Value vs. Coverage for Gridworld (left: Standard Task, middle: Transfer Task), Bottom row: Normalized Value vs. Coverage for Randomworlds (left: Standard Task, middle: Transfer Task). Rightmost plots: Normalized Value vs. Bayesian Regret of both Tasks (top: Gridworld, bottom: Randomworlds).}
    \label{fig:combined_plots_20}
\end{figure*}
\begin{table*}[ht]
\centering
\caption{Comparison of Gridworld and Randomworlds ($20\%$ stochastic-policy states, $\epsilon$ Gridworld = 0.3, $\epsilon$s RandomWorlds = [0.79, 1.11, 0.32, 0.63, 1.16, 0.33, 0.57, 0.47, 0.99, 0.32])}
\label{tab:combined_results_20} 
\resizebox{0.9\textwidth}{!}{ 
\begin{tabular}{l*{5}{r}}
\toprule
\multicolumn{1}{c}{} & \multicolumn{5}{c}{Gridworld} \\
\cmidrule(lr){2-6}
Method & MCE & BITL & ITL & MLE & PS \\
\midrule
$\epsilon$ matching   & 0.64 ± 0.11 & \textbf{0.74 ± 0.08} & 0.64 ± 0.1 & 0.37 ± 0.06 & 0.32 ± 0.01 \\
$\epsilon$ matching transfer & 0.41 ± 0.06 & \textbf{0.51 ± 0.05} & 0.4 ± 0.06 & 0.32 ± 0.04 & 0.32 ± 0.01 \\
Best matching & 0.59 ± 0.11 & \textbf{0.69 ± 0.08} & 0.6 ± 0.1 & 0.34 ± 0.06 & 0.29 ± 0.02 \\
Best matching transfer & 0.4 ± 0.06 & \textbf{0.5 ± 0.05} & 0.39 ± 0.06 & 0.3 ± 0.04 & 0.31 ± 0.02 \\
Time & 114.66 ± 27.54 & 14267.26 ± 529.69 & 0.75 ± 0.53 & \textbf{0.01 ± 0.00}  & 7.82 ± 0.32\\
Total Variation & \textbf{137.81 ± 3.79} & 160.22 ± 3.89 & 138.71 ± 3.5 & 141.67 ± 3.11 & 160.4 ± 3.62 \\
Value CVaR 1\% & 17.94 ± 7.28 & \textbf{103.22 ± 7.96} & 102.83 ± 33.76 & -5.23 ± 0.56 & -3.49 ± 0.56 \\
Value CVaR 2\% & 54.96 ± 16.37 & \textbf{106.6 ± 7.27} & 104.7 ± 25.52 & -5.04 ± 0.57 & -2.9 ± 0.63 \\
Value CVaR 5\% & 106.58 ± 42.1 & \textbf{109.18 ± 5.81} & 107.58 ± 16.89 & -4.74 ± 0.5 & -1.65 ± 0.92 \\
Nbr constraints violated & 0.82 ± 2.53 & \textbf{0.0 ± 0.0} & \textbf{0.0 ± 0.0} & 20.6 ± 5.73 & 14.19 ± 2.53 \\
\midrule
\multicolumn{1}{c}{} & \multicolumn{5}{c}{Randomworlds} \\
\cmidrule(lr){2-6}
Method & MCE & BITL & ITL & MLE & PS \\
\midrule
$\epsilon$ matching & 0.5 ± 0.13 & 0.72 ± 0.13 & \textbf{0.73 ± 0.15} & 0.38 ± 0.13 & 0.24 ± 0.05 \\
$\epsilon$ matching transfer & 0.42 ± 0.12 & 0.34 ± 0.11 & \textbf{0.47 ± 0.13} & 0.42 ± 0.12 & 0.25 ± 0.05 \\
Best matching & 0.45 ± 0.12 & 0.66 ± 0.13 & \textbf{0.68 ± 0.15} & 0.34 ± 0.12 & 0.21 ± 0.05 \\
Best matching transfer & 0.39 ± 0.13 & 0.31 ± 0.1 & \textbf{0.43 ± 0.14} & 0.39 ± 0.13 & 0.22 ± 0.05 \\
Time & 31.96 ± 25.36 & 5493 ± 271.62 & 0.68 ± 0.69 & \textbf{0.0 ± 0.0} & 2.24 ± 0.28 \\
Total Variation & 112.12 ± 2.43 & 124.38 ± 4.55 & \textbf{103.32 ± 4.91} & 112.07 ± 2.23 & 128.34 ± 2.51 \\
Value CVaR 1\% & -520.09 ± 1.89 & \textbf{-401.79 ± 23.07} & -408.13 ± 0.14 & -523.71 ± 1.51 & -428.06 ± 0.79 \\
Value CVaR 2\% & -519.89 ± 1.72 & \textbf{-388.77 ± 29.86} & -396.53 ± 5.59 & -520.58 ± 1.96 & -426.44 ± 1.25 \\
Value CVaR 5\% & -450.23 ± 25.6 & -356.37 ± 33.32 & \textbf{-352.47 ± 21.57} & -451.59 ± 23.83 & -421.73 ± 2.52 \\
Nbr constraints violated & 7.03 ± 3.94 & \textbf{0.0 ± 0.0} & \textbf{0.0 ± 0.0} & 9.5 ± 3.95 & 9.22 ± 2.52 \\
\bottomrule
\end{tabular} 
}
\end{table*}
\begin{figure*}[ht]
    \centering
    \begin{minipage}[b]{0.9\textwidth}
        \centering
        \includegraphics[width=\textwidth]{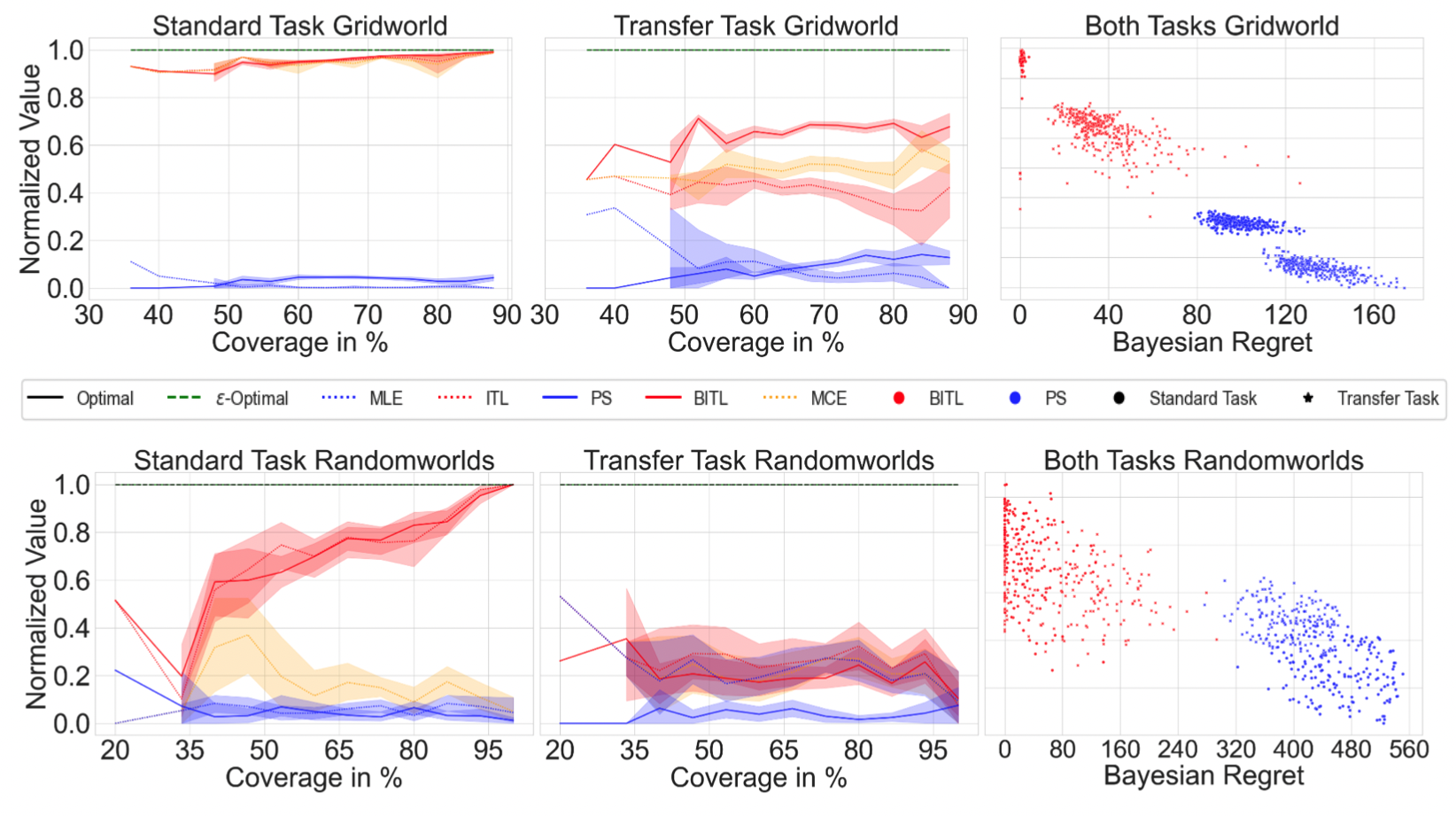}
    \end{minipage}
    \caption{($0\%$ stochastic-policy states, $\epsilon$ Gridworld = 0.0, $\epsilon$s RandomWorlds = 0.0) Top row: Normalized Value vs. Coverage for Gridworld (left: Standard Task, middle: Transfer Task), Bottom row: Normalized Value vs. Coverage for Randomworlds (left: Standard Task, middle: Transfer Task). Rightmost plots: Normalized Value vs. Bayesian Regret of both Tasks (top: Gridworld, bottom: Randomworlds).}
    \label{fig:combined_plots_0}
\end{figure*}
\begin{table*}[ht]
\centering
\caption{Comparison of Gridworld and Randomworlds ($0\%$ stochastic-policy states, $\epsilon$ Gridworld = 0.0, $\epsilon$s RandomWorlds = 0.0)}
\label{tab:combined_results_0} 
\resizebox{0.9\textwidth}{!}{ 
\begin{tabular}{l*{5}{r}}
\toprule
\multicolumn{1}{c}{} & \multicolumn{5}{c}{Gridworld} \\
\cmidrule(lr){2-6}
Method & MCE & BITL & ITL & MLE & PS \\
\midrule
$\epsilon$ matching   & 0.68 ± 0.09 & \textbf{0.75 ± 0.07} & 0.68 ± 0.09 & 0.37 ± 0.06 & 0.29 ± 0.01 \\
$\epsilon$ matching transfer & 0.4 ± 0.06 & \textbf{0.49 ± 0.05} & 0.38 ± 0.06 & 0.29 ± 0.04 & 0.31 ± 0.01 \\
Best matching & 0.68 ± 0.09 & \textbf{0.75 ± 0.07} & 0.68 ± 0.09 & 0.37 ± 0.06 & 0.29 ± 0.01 \\
Best matching transfer & 0.4 ± 0.06 & \textbf{0.49 ± 0.05} & 0.38 ± 0.06 & 0.29 ± 0.04 & 0.31 ± 0.01 \\
Time & 119.86 ± 22.01 & 15284.26 ± 391.76 & 4.36 ± 1.46 & \textbf{0.01 ± 0.0}  & 11.62 ± 0.55 \\
Total Variation & \textbf{137.38 ± 3.3} & 158.41 ± 3.37 & 138.6 ± 3.06 & 141.52 ± 2.55 & 160.22 ± 2.95 \\
Value CVaR 1\% & 75.34 ± 19.3 & \textbf{104.24 ± 7.94} & 104.24 ± 37.65 & -5.23 ± 0.48 & -4.11 ± 0.22 \\
Value CVaR 2\% & 102.59 ± 21.97 & \textbf{108.18 ± 6.74} & 105.64 ± 26.95 & -5.2 ± 0.43 & -3.65 ± 0.34 \\
Value CVaR 5\% & 106.74 ± 23.68 & \textbf{110.83 ± 5.93} & 108.95 ± 18.49 & -4.94 ± 0.35 & -1.88 ± 0.84 \\
Nbr constraints violated & 0.08 ± 0.59 & \textbf{0.0 ± 0.0} & \textbf{0.0 ± 0.0} & 22.4 ± 5.23 & 13.15 ± 1.83 \\
\midrule
\multicolumn{1}{c}{} & \multicolumn{5}{c}{Randomworlds} \\
\cmidrule(lr){2-6}
Method & MCE & BITL & ITL & MLE & PS \\
\midrule
$\epsilon$ matching & 0.46 ± 0.13 & 0.77 ± 0.14 & \textbf{0.78 ± 0.15} & 0.32 ± 0.13 & 0.19 ± 0.05 \\
$\epsilon$ matching transfer & 0.37 ± 0.14 & 0.34 ± 0.13 & \textbf{0.39 ± 0.16} & 0.36 ± 0.15 & 0.23 ± 0.05 \\
Best matching & 0.46 ± 0.13 & 0.77 ± 0.14 & \textbf{0.78 ± 0.15} & 0.32 ± 0.13 & 0.19 ± 0.05 \\
Best matching transfer & 0.37 ± 0.14 & 0.34 ± 0.13 & \textbf{0.39 ± 0.16} & 0.36 ± 0.15 & 0.23 ± 0.05 \\
Time & 32.11 ± 27.16 & 5493 ± 216.83 & 0.91 ± 0.55 & \textbf{0.0 ± 0.0} & 2.35 ± 0.46 \\
Total Variation & 111.47 ± 2.27 & 117.66 ± 11.76 & \textbf{101.44 ± 2.87} & 111.84 ± 2.26 & 128.14 ± 2.54 \\
Value CVaR 1\% & -520.09 ± 3.18 & \textbf{-343.11 ± 0.56} & -350.2 ± 11.25 & -521.78 ± 0.0 & -428.28 ± 0.05 \\
Value CVaR 2\% & -517.71 ± 2.71 & \textbf{-278.91 ± 36.29} & -343.9 ± 20.59 & -520.09 ± 2.39 & -427.96 ± 0.4 \\
Value CVaR 5\% & -446.29 ± 4.08 & \textbf{-246.36 ± 41.54} & -254.84 ± 47.72 & -447.26 ± 3.11 & -425.21 ± 1.24 \\
Nbr constraints violated & 4.9 ± 1.71 & \textbf{0.0 ± 0.0} & \textbf{0.0 ± 0.0} & 6.9 ± 1.91 & 8.59 ± 2.03 \\
\bottomrule
\end{tabular} 
}
\end{table*}

\begin{table*}[ht]
    \centering
    \caption{Healthcare dataset results ($\epsilon = 10$)}
    \resizebox{0.90\textwidth}{!}{ 
    \begin{tabular}{l*{10}{r}}
    \toprule
    \multicolumn{1}{c}{} & \multicolumn{5}{c}{Standard Task} & \multicolumn{5}{c}{Transfer Task} \\
    \cmidrule(lr){2-6} \cmidrule(lr){7-11}
    Method & \textbf{MLE} & ITL & MCE & BITL & PS & MLE & ITL & MCE & BITL & PS \\
    \midrule
    Best matching & 0.33 & \textbf{0.47} & 0.38 & 0.46 & 0.31 & 0.34 & \textbf{0.50} & 0.10 & 0.47 & 0.34 \\
    $\epsilon$ matching   & 0.52 & \textbf{1} & 0.68 & \textbf{1} & 0.51 & 0.58 & \textbf{0.94} & 0.16  & \textbf{0.94} & 0.58 \\
    Nbr Constraints & 49 & \textbf{0} & 43   & \textbf{0} & 52 & - & - & - & - & - \\
    Time & - & \textbf{2.31} & 180 & - & - & - & - & - & - & - \\
    Bayesian Regret & - & - & - & \textbf{0.55} & 10 & - & - & - & \textbf{0.44} & 5.40 \\
    \bottomrule
    \end{tabular}
    }
\label{tab:method_comparison_hypo_10}
\end{table*}
\begin{table*}[ht]
    \centering
    \caption{Healthcare dataset results ($\epsilon = 15$)}
    \resizebox{0.90\textwidth}{!}{ 
    \begin{tabular}{l*{10}{r}}
    \toprule
    \multicolumn{1}{c}{} & \multicolumn{5}{c}{Standard Task} & \multicolumn{5}{c}{Transfer Task} \\
    \cmidrule(lr){2-6} \cmidrule(lr){7-11}
    Method & \textbf{MLE} & ITL & MCE & BITL & PS & MLE & ITL & MCE & BITL & PS \\
    \midrule
    Best matching & 0.33 & \textbf{0.48} & 0.25 & 0.49 & 0.31 & 0.34 & \textbf{0.47} & 0.15 & \textbf{0.47} & 0.34 \\
    $\epsilon$ matching   & 0.52 & \textbf{1} & 0.43 & \textbf{1} & 0.51 & 0.58 & \textbf{0.97} & 0.31  & \textbf{0.97} & 0.58 \\
    Nbr Constraints & 49 & \textbf{0} & 51   & \textbf{0} & 52 & - & - & - & - & - \\
    Time & - & \textbf{1.78} & 94 & - & - & - & - & - & - & - \\
    Bayesian Regret & - & - & - & \textbf{1.02} & 10 & - & - & - & \textbf{0.56} & 5.40 \\
    \bottomrule
    \end{tabular}
    }
    \label{tab:method_comparison_hypo_15}
\end{table*}



\end{document}